%% file: 3DSES__an_indoor_Lidar_point_cloud_segmentation_dataset_with_real_and_pseudo_labels_from_a_3D_model.tex
\documentclass[a4paper,twoside]{article}

\usepackage{epsfig}
\usepackage{subcaption}
\usepackage{calc}
\usepackage{amssymb}
\usepackage{amstext}
\usepackage{amsmath}
\usepackage{amsthm}
\usepackage{multicol}
\usepackage{pslatex}
\usepackage{apalike}
\usepackage[ruled,vlined]{algorithm2e}
\usepackage[bottom]{footmisc}
\usepackage{censor}
\usepackage{siunitx}
\usepackage[breaklinks,colorlinks,citecolor=cvprblue]{hyperref}

\usepackage{tikz}
\usepackage{pgfplots}
\pgfplotsset{compat=1.17}
\usepackage{iftex}
\ifLuaTeX
    \usepackage{emoji}
    \usepackage{fontspec}
\else
    \usepackage{mathptmx}
    \newcommand\emoji[1]{\includegraphics[height=1em]{emojis/#1.png}}
\fi

\usepackage[group-separator={\ },detect-all]{siunitx}  %
\usepackage{amssymb}                        %
\usepackage{pifont}%
\newcommand{\cmark}{\emoji{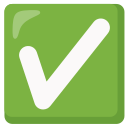}}
\newcommand{\xmark}{\emoji{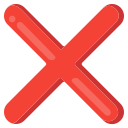}}
\usepackage{float}                          %

\definecolor{cvprblue}{rgb}{0.21,0.49,0.74}
\usepackage{stmaryrd}
\usepackage{subcaption}
\usepackage{booktabs}
\usepackage{tabularx}
\newcolumntype{Y}{>{\centering\arraybackslash}X}
\usepackage{microtype}
\usepackage{xspace}
\usepackage{multirow}
\usepackage{todonotes}

\usepackage{changepage}  %
\usepackage{censor}
\usepackage[capitalize]{cleveref}

\usepackage{SCITEPRESS}     %

\newcommand\etc{etc.\xspace}
\newcommand\ie{\emph{i.e.}\xspace}
\newcommand\eg{\emph{e.g.}\xspace}

\renewcommand\paragraph[1]{\vspace{1ex}\textbf{#1}}

\StopCensoring

\begin{document}

\title{3DSES: an indoor Lidar point cloud segmentation dataset with real and pseudo-labels from a 3D model}

\author{\authorname{Maxime Mérizette\sup{1,2,4}, 
                    Nicolas Audebert\sup{3,4},
                    Pierre Kervella\sup{1,2},
                    Jérôme Verdun \sup{2}}
\affiliation{\sup{1}QUARTA, F-35136 Saint Jacques de la Lande, France}
\affiliation{\sup{2}Conservatoire national des arts et métiers, GeF, EA4630, F-72000 Le Mans, France}
\affiliation{\sup{3} Univ. Gustave Eiffel, ENSG, IGN, LASTIG, F-94160 Saint-Mandé, France}
\affiliation{\sup{4}Conservatoire national des arts et métiers, CEDRIC, EA4629, F-75141 Paris, France}
\email{\{maxime.merizette, jerome.verdun\}@lecnam.net, nicolas.audebert@ign.fr, p.kervella@quarta.fr}
}


\abstract{\input{sec/0_abstract}}

\onecolumn \maketitle \normalsize \setcounter{footnote}{0} \vfill

\section{\uppercase{Introduction}}
\label{sec:introduction}

\input{sec/1_intro}

\section{\uppercase{Previous work}}
\label{sec:related}

\input{sec/2_Related_work}

\section{\uppercase{3DSES}}
\label{3dses}

    \input{sec/3_3DSES}

\section{\uppercase{Experiments}}
\label{experiments}

\input{sec/4_Experiments}

\section{\uppercase{Conclusion}}
\label{conclusion}

\input{sec/5_Conclusion}

\section*{\uppercase{Acknowledgements}}

We would like to express our sincere appreciation to all individuals and organizations who contributed to our paper.
Special thanks to Leica Geosystems for loaning the RTC360 used in the acquisitions.
We acknowledge the support \censor{ESGT} by loaning the Trimble X7 and their permissions to carry out and publish the 3D scans.
We also extend our thanks to \censor{Lilian Ribet} for 3D acquisitions and to \censor{Léa Corduri}, \censor{Judicaëlle Djeudji Tchaptchet}, \censor{Damien Richard} and their supervisor Élisabeth Simonetto for 3D manual annotations.

\bibliographystyle{apalike}
{\small
\bibliography{bibliography}}

\section*{\uppercase{Appendix}}

\input{sec/X_suppl}

    \input{sec/X_suppl_datasheet}

\end{document}

%% file: sec/0_abstract.tex
    Semantic segmentation of indoor point clouds has found various applications in the creation of digital twins for robotics, navigation and building information modeling (BIM). However, most existing datasets of labeled indoor point clouds have been acquired by photogrammetry.
     In contrast, Terrestrial Laser Scanning (TLS) can acquire dense sub-centimeter point clouds and has become the standard for surveyors. 
     We present 3DSES (3D Segmentation of \censor{ESGT} point clouds), a new dataset of indoor dense TLS colorized point clouds covering \SI{427}{\meter\squared} of an engineering school.
     3DSES has a unique double annotation format: semantic labels annotated at the point level alongside a full 3D CAD model of the building. 
     We introduce a model-to-cloud algorithm for automated labeling of indoor point clouds using an existing 3D CAD model.
     3DSES has 3 variants of various semantic and geometrical complexities.
     We show that our model-to-cloud alignment can produce pseudo-labels on  our point clouds with a $>95\%$ accuracy, allowing us to train deep models with significant time savings compared to manual labeling.
      First baselines on 3DSES show the difficulties encountered by existing models when segmenting objects relevant to BIM, such as light and safety utilities.
      We show that segmentation accuracy can be improved by leveraging pseudo-labels and Lidar intensity, an information rarely considered in current datasets.
     Code and data is open sourced.

%% file: sec/1_intro.tex
\input{Image/Figure1bis}

\input{Table/Dataset_RelatedWork}

    Building Information Modeling (BIM) is a comprehensive tool for managing buildings throughout their entire life cycle, from construction to demolition. It consists in creating a digital representation of a building, called a ``digital twin''. BIM helps reduce construction and maintenance costs by facilitating planning and simulation on the virtual assets \cite{BRADLEY2016139} and preserve heritage structures \cite{pocobelli2018bim}. BIM allows for monitoring buildings over time and managing equipment by recording details such as installation date and maintenance schedules.
    The creation of digital twins often involves \emph{in situ} acquisitions to reconstruct the building's 3D structure, often using point clouds \cite{wang2015integrating,JUNG2018811,isprs-archives-XLII-5-W1-47-2017}.
    In recent years, 3D data acquisition technologies have not only significantly improved in accuracy, but also diversified their sensing apparatus. In most cases, sensors create point clouds based either on photogrammetry, \eg using stereo photography or structure-from-motion, or on laser-based Lidar systems. Acquisition has been made increasingly intuitive and easy with the improvements of 3D scanners, including real-time positioning and very high acquisition speed. Terrestrial Laser Scanning (TLS) has become the standard for surveyors to create large point clouds of building interiors in a few hours.
    
    Meanwhile, the enrichment of point clouds has not met the same progresses. 3D CAD modeling of buildings based on point clouds remains a manual and time-consuming task. Creation of 3D CAD models is minimally automated and still requires the intervention of qualified experts.
    Semantic segmentation of point clouds is a promising avenue to automatically label point clouds, and could accelerate the modeling by helping surveyors to identify structural primitives (walls, ground, doors) and even furniture types (chairs, tables, \etc).
    However, few datasets exist for semantic segmentation of indoor TLS point clouds.
    Moreover, surveying companies have access to large databases of existing 3D CAD models and associated point clouds, but the latter are mostly unlabeled.
    For these reasons, we introduce 3DSES (\cref{fig:main}), a dataset of indoor TLS acquisitions with manually annotated point clouds and a BIM-like 3D CAD model.
    In addition to the overall structure and furniture, we label several types of common BIM elements, such as extinguishers, alarms and lights, that are challenging to detect in point clouds.
    To evaluate the feasibility of automatically annotating point clouds based on existing BIM models, we introduce a 3D model-to-cloud alignment algorithm to label points clouds.
    We show that these pseudo-labels are nearly as effective as manual point cloud annotation for most classes. However, we show that small objects remain extremely challenging for existing point cloud segmentation models.
    3DSES is a unique dataset that contains all the steps required for automated scan-to-BIM: dense point clouds, semantic segmentation labels and a full 3D CAD model. We hope that 3DSES will enable the creation and testing of deep models for multiple tasks, from point cloud segmentation to BIM generation through mesh to point cloud alignment.

%% file: Image/Figure1bis.tex
\definecolor{column_color}{rgb}{0.51, 0.04, 0.46}  %
\definecolor{components_color}{rgb}{0.16,0.56,0.03}%
\definecolor{covering_color}{rgb}{0.71,0.98,0.62}  %
\definecolor{damper_color}{rgb}{0.02,1., 0.95}     %
\definecolor{door_color}{rgb}{0.22,0.6,0.58}       %
\definecolor{lightfixture_color}{rgb}{0.65,0.96,0.95}%
\definecolor{fire_color}{rgb}{0.12,0.02,0.99}      %
\definecolor{furniture_color}{rgb}{0.14,0.09,0.59} %
\definecolor{heater_color}{rgb}{0.54,0.49,0.94}    %
\definecolor{lamp_color}{rgb}{0.46,0.73,0.8}       %
\definecolor{outlet_color}{rgb}{0.93,0.73,0.91}    %
\definecolor{railing_color}{rgb}{0.97,0.07,0.03}   %
\definecolor{slab_color}{rgb}{1.,0.89,0.}          %
\definecolor{stair_color}{rgb}{0.97,0.66,0.64}     %
\definecolor{switch_color}{rgb}{0.75,0.19,0.16}    %
\definecolor{wall_color}{rgb}{0.79,0.77,0.69}      %
\definecolor{window_color}{rgb}{1.,0.96,0.69}      %
\definecolor{clutter_color}{rgb}{1.,0.01,0.}       %

\begin{figure}[ht]
    \centering
    \begin{subfigure}[b]{0.24\textwidth}
        \centering
        \includegraphics[width=\textwidth]{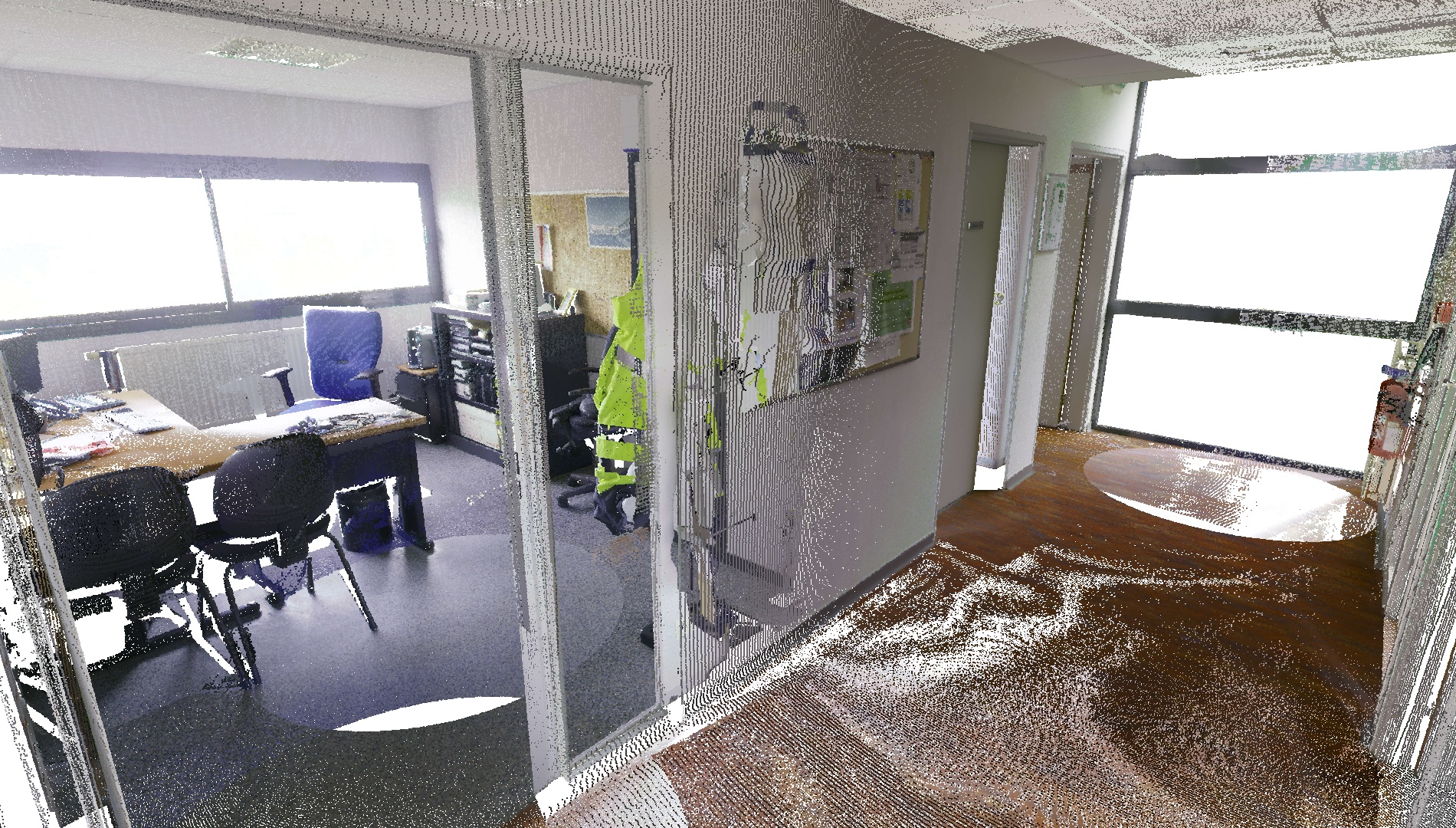}
        \caption{RGB}\vspace{5pt}
        \label{fig:rgb}
    \end{subfigure}%
    \begin{subfigure}[b]{0.24\textwidth}
        \centering
        \includegraphics[width=\textwidth]{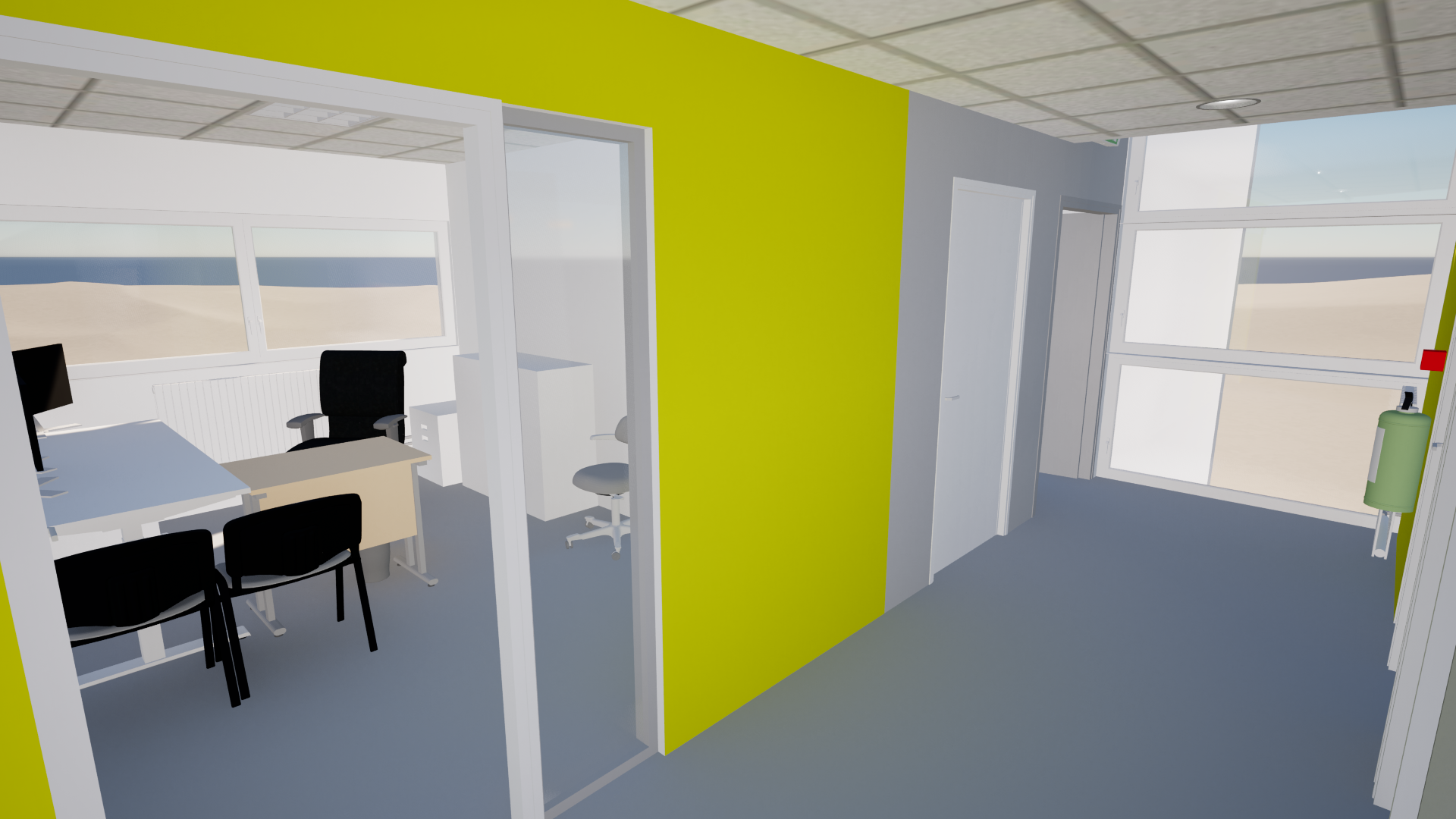}
        \caption{3D model}\vspace{5pt}
        \label{fig:3dmodel}
    \end{subfigure}
    
    \begin{subfigure}[b]{0.24\textwidth}
        \centering
        \includegraphics[width=\textwidth]{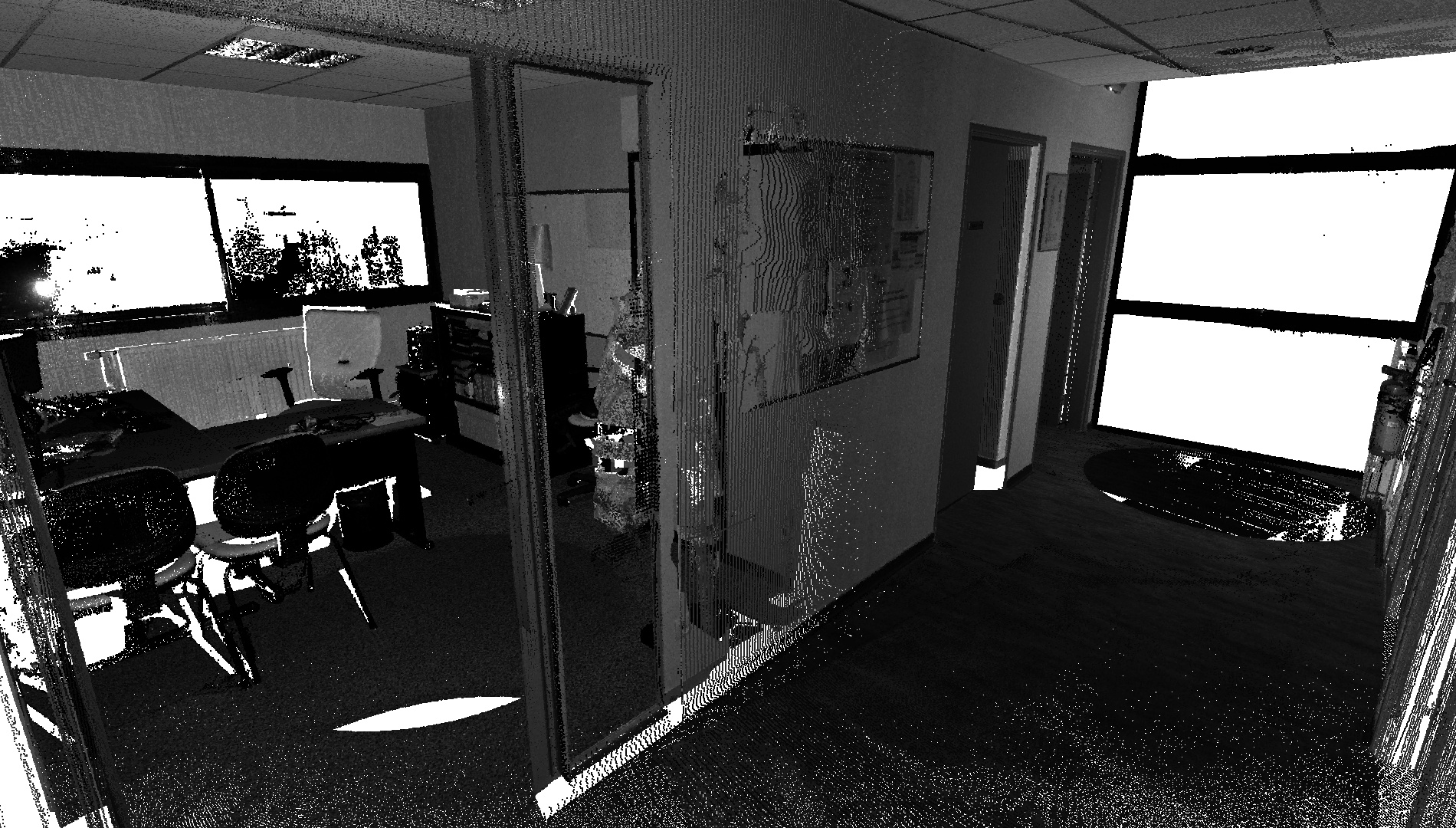}
        \caption{Intensity}\vspace{5pt}
        \label{fig:intensity}
    \end{subfigure}%
    \begin{subfigure}[b]{0.24\textwidth}
        \centering
        \includegraphics[width=\textwidth]{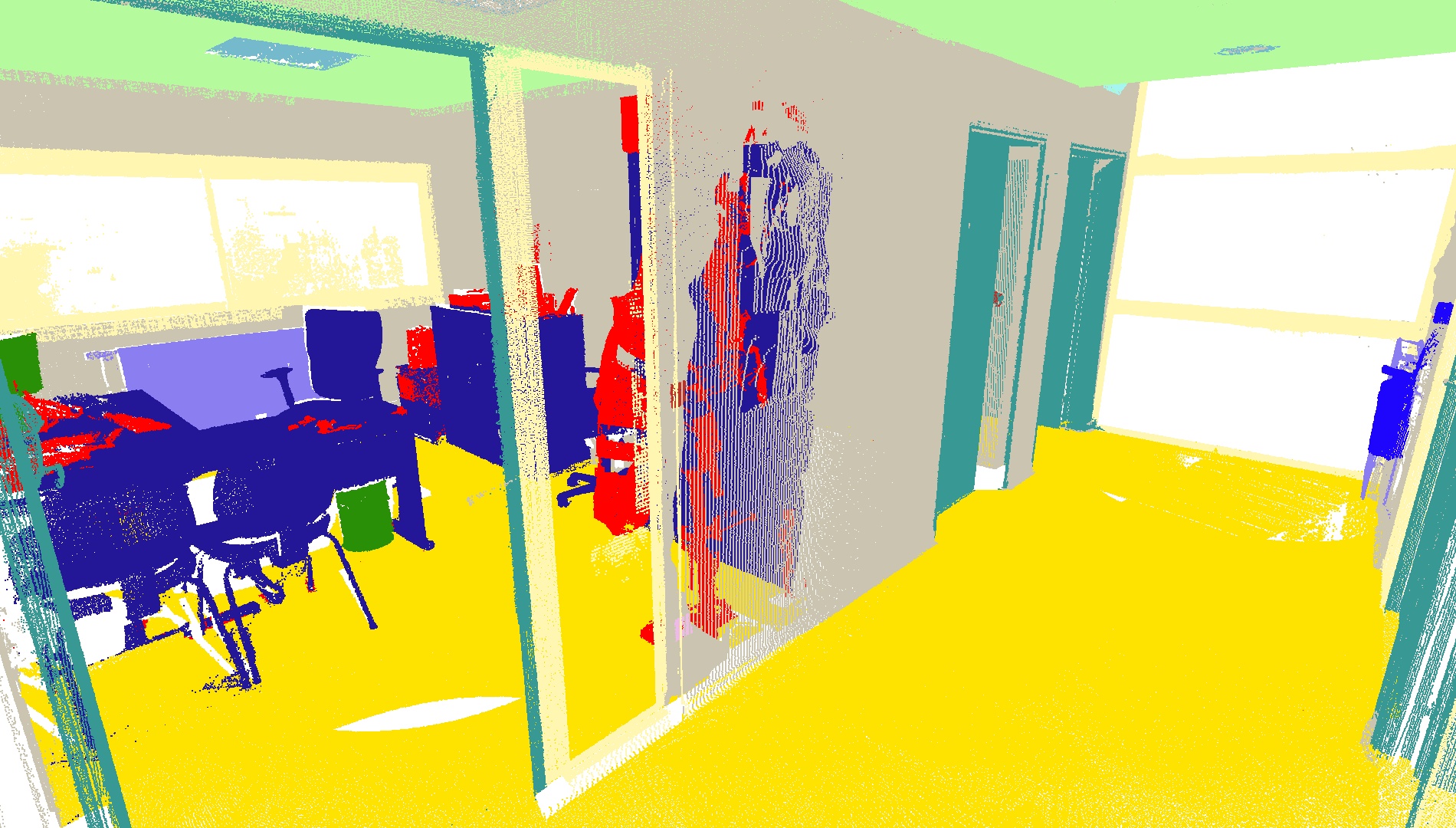}
        \caption{Gold - Real labels}\vspace{5pt}
        \label{fig:goldreal}
    \end{subfigure}
    
    \begin{subfigure}[b]{0.24\textwidth}
        \centering
        \includegraphics[width=\textwidth]{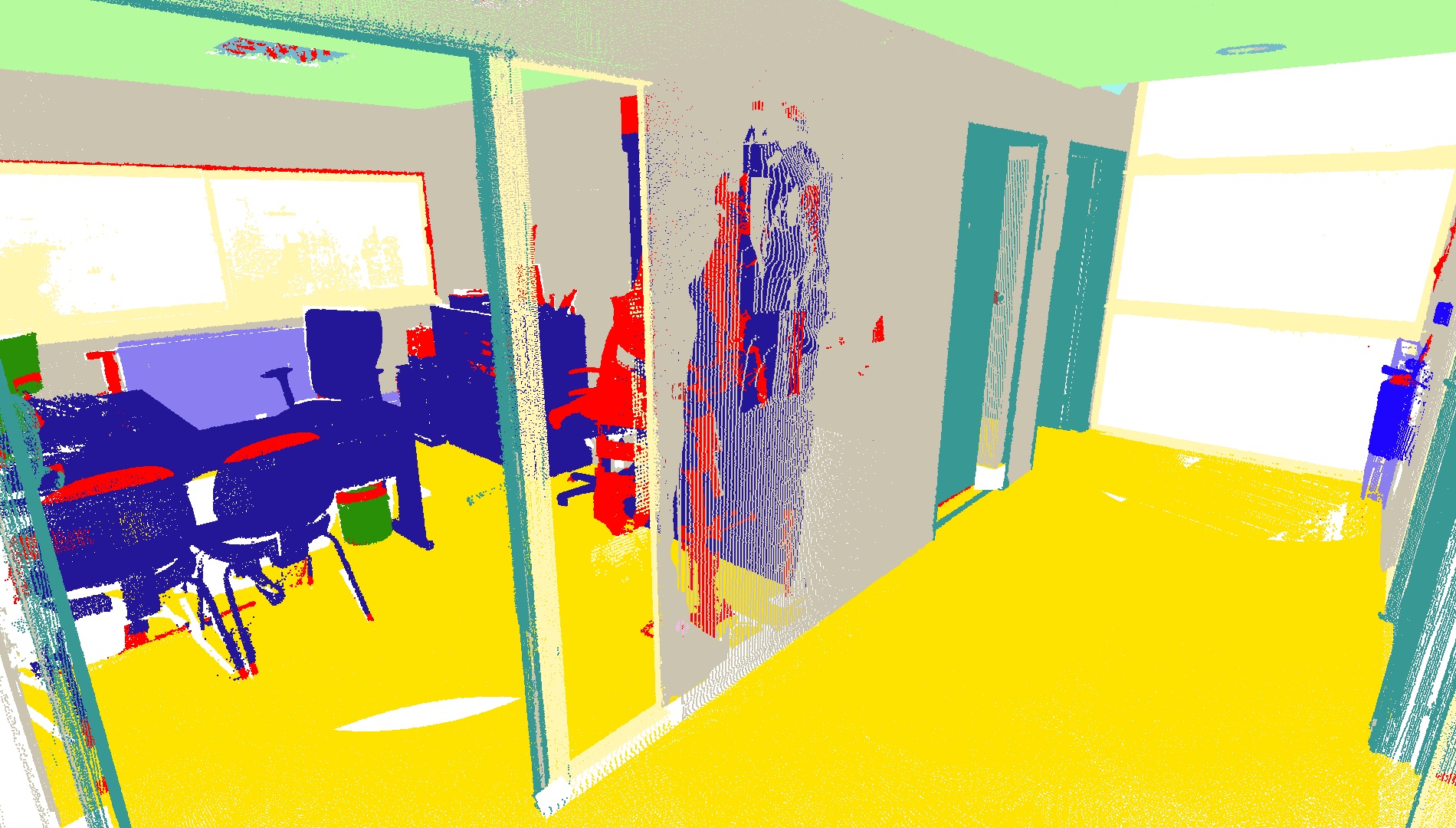}
        \caption{Gold - Pseudo labels }\vspace{5pt}
        \label{fig:goldpseudo}
    \end{subfigure}%
    \begin{subfigure}[b]{0.24\textwidth}
        \centering
        \includegraphics[width=\textwidth]{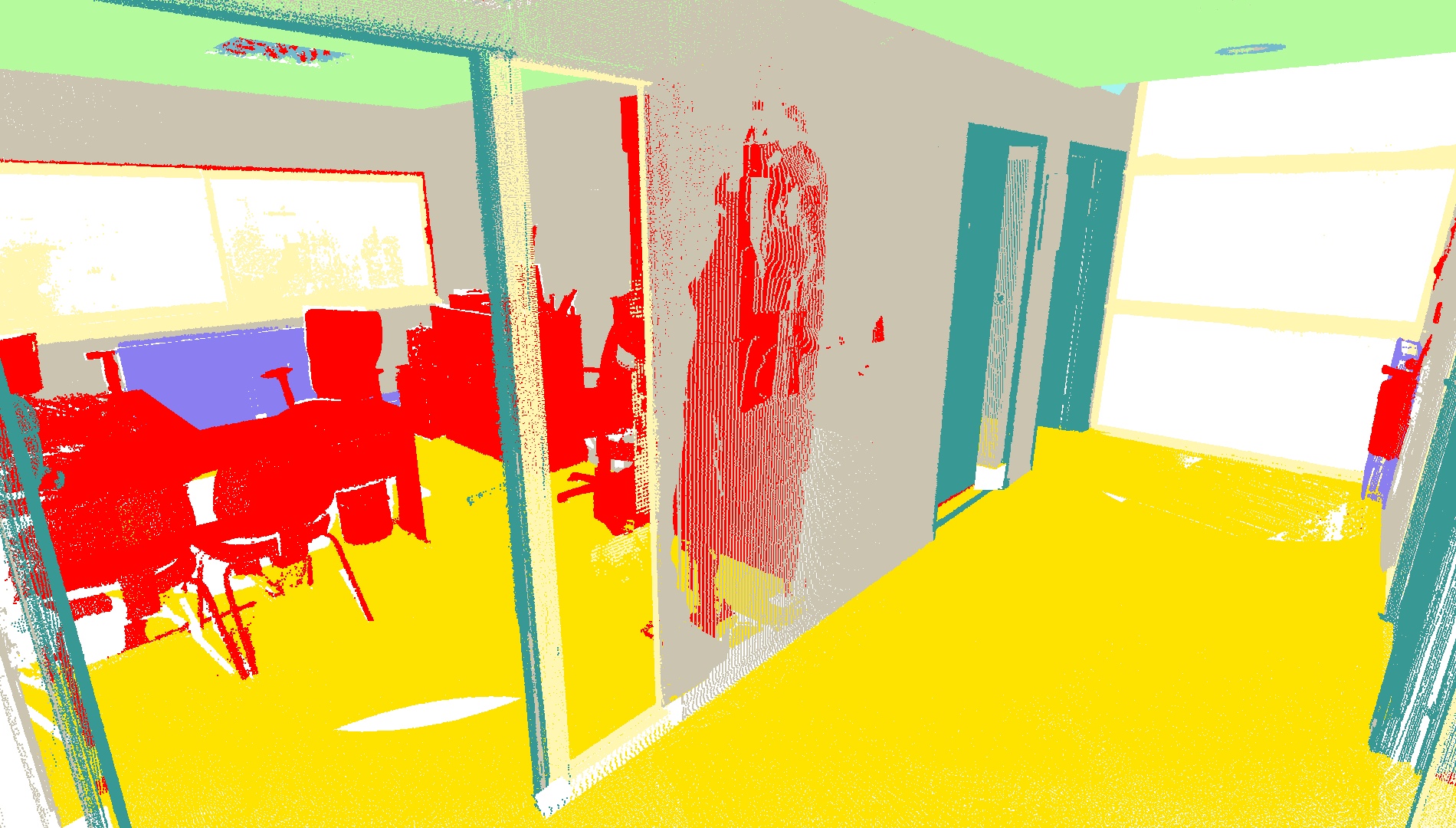}
        \caption{Silver - Pseudo labels}\vspace{5pt}
        \label{fig:silverpseudo}
    \end{subfigure}
    
    \caption{Modalities and annotation variants of 3DSES. Gold real labels are manual annotations across 18 classes, including small objects such as light switches and electrical outlets. Pseudo-labels are obtained by automatically aligning the 3D model on the point cloud, introducing some noise in the annotation (see \eg the top of the chairs). Silver labels use a simplified classification of only 12 categories (\eg the wastebin is now simply ``clutter''). Legend: Column in \textcolor{column_color}{dark purple},
    components in \textcolor{components_color!50!black}{dark green},
    coverings in \textcolor{covering_color!50!black}{light green},
    doors in \textcolor{door_color}{green},
    emergency signs in \textcolor{lightfixture_color}{light blue},
    fire terminals in \textcolor{fire_color}{dark blue}, 
    heaters in \textcolor{heater_color}{light purple},
    lamps in \textcolor{lamp_color}{blue}, 
    ground in \textcolor{slab_color!70!black}{yellow},
    walls in \textcolor{wall_color}{grey},
    windows in \textcolor{window_color!70!orange}{light yellow},
    clutter in \textcolor{clutter_color}{red}.}
    \label{fig:main}
\end{figure}

%% file: Table/Dataset_RelatedWork.tex
\begin{table*}[t]
\caption{Comparison of the characteristics of various point cloud datasets from the literature. Note that 3DSES is the only indoor TLS dataset that includes intensity, point level annotations and a 3D CAD model. Despite its size, it also has more points than most existing datasets, demonstrating a very high point density.}
\label{tab:datasets}

\setlength{\tabcolsep}{2pt}
    \begin{tabularx}{\textwidth}{l>{\small}cYccYY>{\small}c}
    \toprule
    \textbf{\small Name} & 
    \textbf{\small Environment} &
    \textbf{\small Classes} &
    \textbf{\small Extent}\textsuperscript{1} & 
    \textbf{\small Points (M)} &
    \textbf{\small Intensity} & 
    \textbf{\small 3D model} & 
    \textbf{\small Source} \\
    \midrule

    Oakland \tiny \cite{munoz_contextual_2009} & Outdoor & 44 & - & 1.6 & \xmark & \xmark & MLS \\
    Paris-rue-Madame \tiny \cite{noauthor_paris-rue-madame_2014} & Outdoor & 17 & \SI{160}{\meter} & 20 & \cmark & \xmark & MLS \\
    IQmulus \tiny \cite{vallet_terramobilitaiqmulus_2015} & Outdoor & 8 & \SI{10000}{\meter} & 12 & \cmark & \xmark & MLS \\
    Semantic 3D \tiny \cite{hackel_semantic3dnet_2017} & Outdoor & 8 & - & 4000 & \cmark & \xmark & TLS \\
    Paris-Lille-3D \tiny \cite{roynard_paris-lille-3d_2018} & Outdoor & 9  & \SI{1940}{\meter} & 143.1 & \cmark & \xmark & MLS \\
    SemanticKITTI \tiny \cite{behley_towards_2021} & Outdoor & 25 & \SI{39200}{\meter} & 4500 & \cmark & \xmark & MLS \\
    Toronto-3D \tiny \cite{tan_toronto-3d_2020} & Outdoor & 8 & \SI{1000}{\meter} & 78.3 & \cmark & \xmark & TLS \\
    \midrule
    
    Matterport3D \tiny \cite{chang_matterport3d_2017} & Indoor & 20 & \SI{219399}{\square\meter} & - & \xmark & \xmark & Camera  \\
    ScanNet \tiny \cite{Dai_2017_CVPR} & Indoor & 20 & \SI{78595}{\square\meter} & 242 & \xmark & \xmark & Camera \\
    S3DIS \tiny \cite{armeni_cvpr16}
     & Indoor & 13 & \SI{6020}{\square\meter} & 215 & \xmark & \xmark & Camera \\
    ScanNet++ \tiny \cite{yeshwanth_scannet_nodate} & Indoor & - & \SI{15000}{\square\meter} & 20 & \xmark & \xmark & TLS \\
    ScanNet200 \tiny \cite{rozenberszki_scannet200_2022} & Indoor & 200 & \SI{78595}{\square\meter} & 242 & \xmark & \xmark & Camera \\
    LiDAR-Net \tiny \cite{guo_lidar-net_2024} & Indoor & 24 & \SI{30000} {\square\meter} & 3600 & \cmark & \xmark & MLS \\
    \textbf{3DSES \- Gold} \emoji{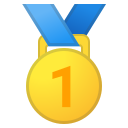} & Indoor & 18 & \SI{101}{\square\meter} & 65 & \cmark & \cmark & TLS \\
    \textbf{3DSES \- Silver} \emoji{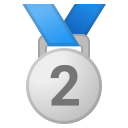} & Indoor & 12 & \SI{304}{\square\meter} & 216 & \cmark & \cmark & TLS \\
    \textbf{3DSES \- Bronze} \emoji{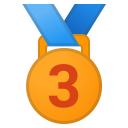} & Indoor & 12 & \SI{427}{\square\meter} & 413 & \cmark & \cmark & TLS \\
    \midrule

    Indoor Modelling \tiny \cite{khoshelham_isprs_2017} & Indoor & \xmark & \SI{2824}{\square\meter} & 127 & \xmark & \cmark  & 5 sensor\\
    Craslab \tiny \cite{abreu_labelled_2023} & Indoor & \xmark & \SI{417}{\square\meter} & 584 & \cmark & \cmark  & TLS \\ 
    \bottomrule

    \end{tabularx}

    \medskip
    {\footnotesize \textsuperscript{1} Surface for indoor datasets, linear extent for outdoor datasets.}
\end{table*}

%% file: sec/2_Related_work.tex
    Numerous datasets exist for semantic segmentation of point clouds with various sizes of scenes, different types of objects of interest and acquired using various sensors, each with their own characteristics. We review in \cref{tab:datasets} some of the more popular ones.
    
    \paragraph{Outdoor datasets} The first popular datasets for semantic segmentation of point clouds focused on outdoors. Mobile laser scanning is popular for outdoor scenes as moving platforms cover more ground. Since the laser is moving, the point clouds tend to be sparse, \eg the seminal Oakland dataset \cite{munoz_contextual_2009} has less than 2M points.
    Later datasets such as IQmulus \cite{vallet_terramobilitaiqmulus_2015} or Paris-rue Madame \cite{noauthor_paris-rue-madame_2014} are also relatively small, with less than 20M points. Bigger datasets have been consolidated by covering larger scenes, such as Paris-Lille-3D \cite{roynard_paris-lille-3d_2018} and SemanticKITTI \cite{behley_towards_2021}. While MLS makes sense for autonomous driving, segmentation performance on these point clouds is not representative of indoor scenes which are much denser with lots of small objects. Concurrently, point clouds acquired by aerial Lidar have been used to create datasets on very large scenes, such as the ISPRS 3D Vaihingen \cite{rottensteiner_isprs_2012}, DublinCity \cite{zolanvari_dublincity_2019}, LASDU \cite{ye_lasdu_2020}, DALES \cite{varney_dales_2020}, Campus3D \cite{Li2020Campus3D}, Hessigheim \cite{kolle_hessigheim_2021}, SensatUrban \cite{hu_towards_2021} and FRACTAL \cite{gaydon_fractal_2024}. These datasets use Aerial Laser Scanning (ALS), with a top-down view that makes them effective for digital surface models but unsuitable for BIM.

    However, some outdoor datasets have a density and geometry close to those found in BIM. For example, Semantic 3D \cite{hackel_semantic3dnet_2017} and Toronto-3D \cite{tan_toronto-3d_2020} both use TLS with high point density. These outdoor scenes do not contain many small objects, though, as they rarely consider classes smaller than outdoor furniture, \eg benches or trashbins.

    \paragraph{Indoor datasets} Few new indoors datasets have been published in the last five years. The two most widely used datasets -- S3DIS \cite{armeni_cvpr16} and ScanNet \cite{Dai_2017_CVPR} -- were published in 2017. The lesser known Matterport3D \cite{chang_matterport3d_2017} was published in the same year with similar characteristics. ScanNet was updated with more classes in ScanNet200 \cite{rozenberszki_scannet200_2022}, yet using the same point clouds. All these datasets are acquired by RGB-D cameras.
    The resulting point clouds are sparser and more sensitive to occlusions than TLS data.
    For example, S3DIS contains 215 million points, which corresponds to approximately ten stations in a medium-resolution TLS system.
    Yet, these datasets are the most common benchmarks to evaluate deep point cloud segmentation, meaning that new approaches are tested on partially obsolete technology.
    While indoor TLS datasets exist, \eg Indoor Modeling \cite{khoshelham_isprs_2017} and Craslab  \cite{abreu_labelled_2023}, they do not contain semantic labels and only release a simplified CAD model.
    LiDAR-Net \cite{guo_lidar-net_2024} uses a mobile laser scanner (MLS) to create an indoor dataset more suitable for autonomous navigation, resulting in a point cloud that contains scan holes, scan lines and various anomalies that are not shared with TLS scans for building surveys.
    To the best of our knowledge, the only dataset using labeled TLS point clouds is ScanNet++ \cite{yeshwanth_scannet_nodate}.
    However, ScanNet++ used a complex three devices acquisition setup. DSLR images were acquired separately from the scans, and then backprojected to colorize point clouds. This setup is not representative of usual surveys practices. For 3DSES, we use a simpler acquisition workflow, as the RGB information comes directly from the TLS.

    \paragraph{Points clouds with intensity} 
    Lidar intensity measures the strength of the laser impulse returned by a scanned point.
    It is a feature commonly used in outdoor point cloud datasets, especially because infrared is helpful to identify vegetation.
    However, intensity is notably absent from indoor datasets, with the exception of LiDAR-Net \cite{guo_lidar-net_2024}.
    In theory, different materials reflect light differently and these variations impact the measured intensity of the laser echo.
    This information might help deep models to discriminate between objects that have similar geometry, but different natures.
    For this reason, we include the intensity information in our 3DSES dataset.
    
    \paragraph{Uniqueness of 3DSES} %
    While covering a smaller surface than other datasets, 3DSES is extremely dense, with a sub-centimeter resolution. It is also the only TLS dataset with Lidar intensity, an information often removed in publicly available datasets, despite theoretically being a discriminative property of materials.
    3DSES is also a \emph{labeled} dataset, suitable to train or evaluate semantic segmentation algorithms.
    Finally, 3DSES comes with a 3D CAD model designed for BIM.
    This combination is unique across existing datasets, and makes 3DSES suitable to investigate 3D point clouds for indoor building surveys and modeling.

%% file: sec/3_3DSES.tex
\input{Image/Nuage_maquette_les2}

    We present in this section the data acquisition and labeling process, the 3D modeling and an automated pseudo-labeling alignment algorithm.%

    \subsection{Data collection}

        \paragraph{Point clouds acquisition}
            Data acquisition was carried out at \censor{ESGT} using two Terrestrial Laser Scanners (TLS): a Leica RTC360 and a Trimble X7. 
            High-resolution pictures were taken for each scan (15MP for RTC360 and 10MP for Trimble X7). Scans were preregistered during the survey.
            We performed and bundled multiple scans inside every room to capture as many objects as possible. Scans were then merged for registration, and any missing link was manually corrected. Point clouds are georeferenced using coordinates from total stations and GNSS.
            We release both colorized (\cref{fig:rgb}) and intensity (\cref{fig:intensity}) clouds.

        \paragraph{Manual labeling} \label{manual_labelling}
            We manually annotated the point clouds to create a ground truth denoted as the \emph{real labels}, shown in \cref{fig:goldreal}.
            Since this is time-consuming, we annotated only 10 points clouds in 18 fine-grained classes: ``Column'', ``Component'', ``Covering'', ``Damper'', ``Door'', ``Exit sign'', ``Fire terminal'', ``Furniture'', ``Heater'', ``Lamp'', ``Outlet'', ``Railing'', ``Slab'', ``Stair'', ``Switch'', ``Wall'', ``Window'' and a ``Clutter'' class that encompasses all points not belonging to another class. 
            Labels were annotated in two passes: 1) labeling by a single annotator (30 to 40 minutes per scan, depending on the complexity of the point cloud, the number of points and the diversity of represented objects); 2) verification pass by an experienced annotator (20 to 30 minutes per scan). %
            
            We then annotated 20 additional point clouds with a simpler taxonomy of only 12 classes, shown in \cref{fig:silverpseudo}. These labels were annotated in a single pass, as the target objects are less ambiguous with simpler geometries. During this process, the points clouds were partially cleaned of outliers and far away points.

        \paragraph{3D CAD model}
            Each type of object is tagged as a member of the corresponding IFC (Industry Foundation Classes) family. The geometry of structural elements (walls, floors, roofs, \etc) is accurately modeled, \ie shapes and dimensions are modeled as precisely as possible. Furniture, such as tables and chairs, and utilities, such as fire extinguishers and emergency exit signs, use standard models, \eg all chairs use the same mesh (cf. \cref{fig:room_model}).
            This is a common practice in BIM, as defining a separate ``chair'' family for each instance would be too time-consuming. \cref{fig:room_overlay} illustrates how these generic 3D CAD models create slight geometrical discrepancies between the point cloud and the model.
            Finally, a special care is given to doors, that can appear either open or closed in scans.
            We model each door in its correct state depending on its true position in the point cloud.
            Complete modeling took slightly less than 30 hours.

    \input{Table/Dataset_available}

    \subsection{Dataset variants} \label{section_dataset}

        Based on the TLS scans and the manual annotations, we built three versions of the 3DSES dataset (cf. \cref{tab:dataset_details}).
        The Gold version is composed of the 10 scans annotated in 18 classes. We consider it to be the ``gold standard'', using fine-grained high quality real labels. We then extended it into a Silver version that contains all the Gold data and an additional 20 scans. Silver labels use a simplified taxonomy of only 12 classes, that are less time consuming to produce.
        Both Gold and Silver variants of 3DSES are high quality, using a real ground truth and cleaned up point clouds.
        Finally, we deliver a Bronze version that includes 12 more scans.
        Bronze contains the raw point clouds and not the processed and cleaned clouds. These full point clouds are denser and noisier, but also more representative of actual field scans.
        Since the additional point clouds have not been manually labeled, the Bronze dataset uses the automatically generated pseudo-labels based on the 3D model using the procedure detailed in \cref{sec:pseudolabeling}.

        Note that all variants suffer from class imbalance, as shown in \cref{fig:histogram_gold,fig:histogram_silverbronze}. Structural elements are over represented compared to other classes, especially furniture and utilities, that are comprised of smaller objects. This is a well-known issue in indoor datasets, such as S3DIS \cite{armeni_cvpr16}, which has $10\times$ more wall points than window points, and ScanNet200 \cite{rozenberszki_scannet200_2022}, which contains 51 million wall points and only \num{50000} fire extinguisher points.

\input{Histograms/Gold_repartion_classes}

        \paragraph{Train/test split} %
        We define a set Train/Val/Test split with a common test area to all variants, based on 3 scans located in the Gold section (scans S170, S171 and S180).
        It contains $\approx$ 20.7 million points with real ground truth. This allows us to evaluate models on real labels only, whether they have been trained on real or pseudo-labels.
        Ground truth labels on the test set are kept hidden for later use in a \href{https://www.codabench.org/}{Codabench} challenge.
    
    \subsection{Pseudo-labeling from the 3D model}
    \label{sec:pseudolabeling}

    One of our goals is to evaluate the feasibility of using existing 3D CAD models to label automatically point clouds for semantic segmentation. 
    Pseudo-labels could help leverage existing databases of surveyed buildings that have been scanned and modeled, but not annotated at the point level.
    To this end, we design an alignment algorithm to map the 3D model on a point cloud.

    First, we divide our 3D CAD model into objects. This allows us to separate individual instances of walls, heaters, light switches and so on. For each object, we produce the corresponding 3D mesh.
    Since the 3D CAD model and the point cloud are georeferenced, we can compute a mesh-to-cloud distance for every point in the point cloud.
    For each object, we first compute its georeferenced bounding box. Then, we compute the distance for each point inside the bounding box to the mesh of the object using the Metro algorithm \cite{cignoni_metro_1998}, implemented in CloudCompare \cite{girardeau-montaut_detection_2006}.
    All points that are inside the mesh are labeled the same class as the IFC family of the object the mesh is derived from.
    To alleviate for geometrical discrepancies between the mesh and the point cloud, points outside the mesh are assigned to their closest mesh as long as the distance is lower than a predefined threshold.
    We then repeat this process for all objects.
    Remaining points that have not been labeled are classified as ``clutter''. This covers objects that are present in the scan, but have not been modeled, \eg jackets on chairs, books and papers on tables, \etc. 
    The algorithms runs in around 9 hours on CPU to align the full dataset (Bronze). This means the pseudo-labeling process (3D model + alignment) takes $\approx$ 40 hours. In comparison, manual point cloud annotation takes 1 hour per scan on average, \ie would have taken 42 hours for 3DSES Bronze, including quality check.
    While these times are comparable, point clouds are intermediate products in indoor surveys, the end goal of which is almost always the production of a 3D CAD model. This is why we assess whether pointwise labels can be obtained as a ``free'' byproduct, without any additional time dedicated to point annotation.

    \paragraph{Evaluation of the pseudo-labels}\label{Matching_score}
        Since 3DSES also includes real labels, we can evaluate how well the pseudo-labels match the ground truth. To do so, we computed some standard segmentation metrics, \ie Intersection over Union (IoU), mean Accuracy (mAcc) and Overall Accuracy (OA).
        We used different confidence thresholds depending on the object class:
        \begin{itemize}
            \item Gold: \SI{4}{\cm} for all classes, except for ``Door'', ``Furniture'', ``Window'', for which we used \SI{10}{\centi\meter}, due to larger uncertainties when modeling;
            \item Silver and Bronze: \SI{4}{\cm} for all classes, except for ``Door'' (\SI{10}{\cm}) and ``Window'' (\SI{15}{\centi\meter}).
        \end{itemize}

\input{Table/Evaluation_Pseudo_LAbel}

        Metrics are computed between pseudo-labels and the manual ground truth over the full dataset.
        We report the alignment metrics in \cref{tab:matching_performances}. We obtain high-quality pseudo-labels on Gold version with $\approx 70$\% mIoU and 95\% accuracy. Structural classes (``Covering'', ``Slab'', ``Wall'') are very well annotated, with a $>$\SI{90}{\%} score. This is expected as these entities have regular shapes with a fine alignment between the 3D model and the point cloud. %
        The lowest scores are on the ``Outlet'' and ``Switch'' classes, below \SI{50}{\%}.
        
        Alignment on the Silver variant is also satisfactory with $\approx75$\% mIoU and $>96$\% accuracy. Metrics are higher on Silver since it focuses on structural classes that are generally easier to align.
        The IoU for ``Column'' is also the lowest due to the use of a slightly too small column diameter in the CAD model. The second worst score is for ``Window' with \SI{69}{\%}, as Silver contains more window types, including frames that deviate from the CAD model. 
        Finally, metrics on ``Railing'' and ``Stair'' are identical on Gold and Silver, since stairs cover the same area in both datasets.

%% file: Image/Nuage_maquette_les2.tex
\begin{figure}[t]

    \centering
    \begin{subfigure}[b]{0.45\textwidth}
        \centering
        \includegraphics[width=\textwidth]{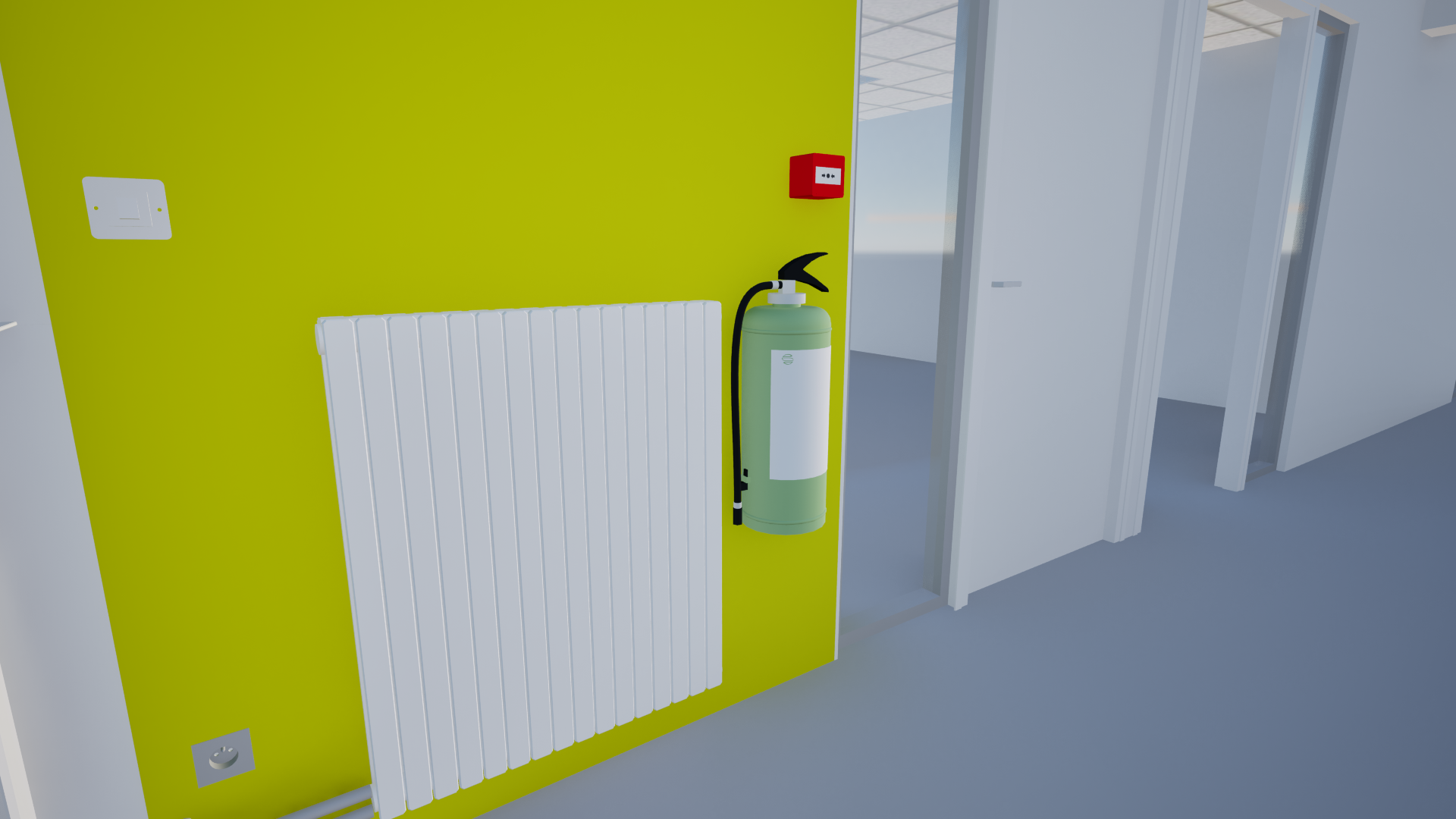}
        \caption{Example of modeled 3D systems: fire alarm, fire extinguisher, heater, outlet, light switch.}\vspace{3pt}
        \label{fig:model_equipment}
    \end{subfigure}
    \hfill
    \begin{subfigure}[b]{0.45\textwidth}
        \centering
        \includegraphics[width=\textwidth]{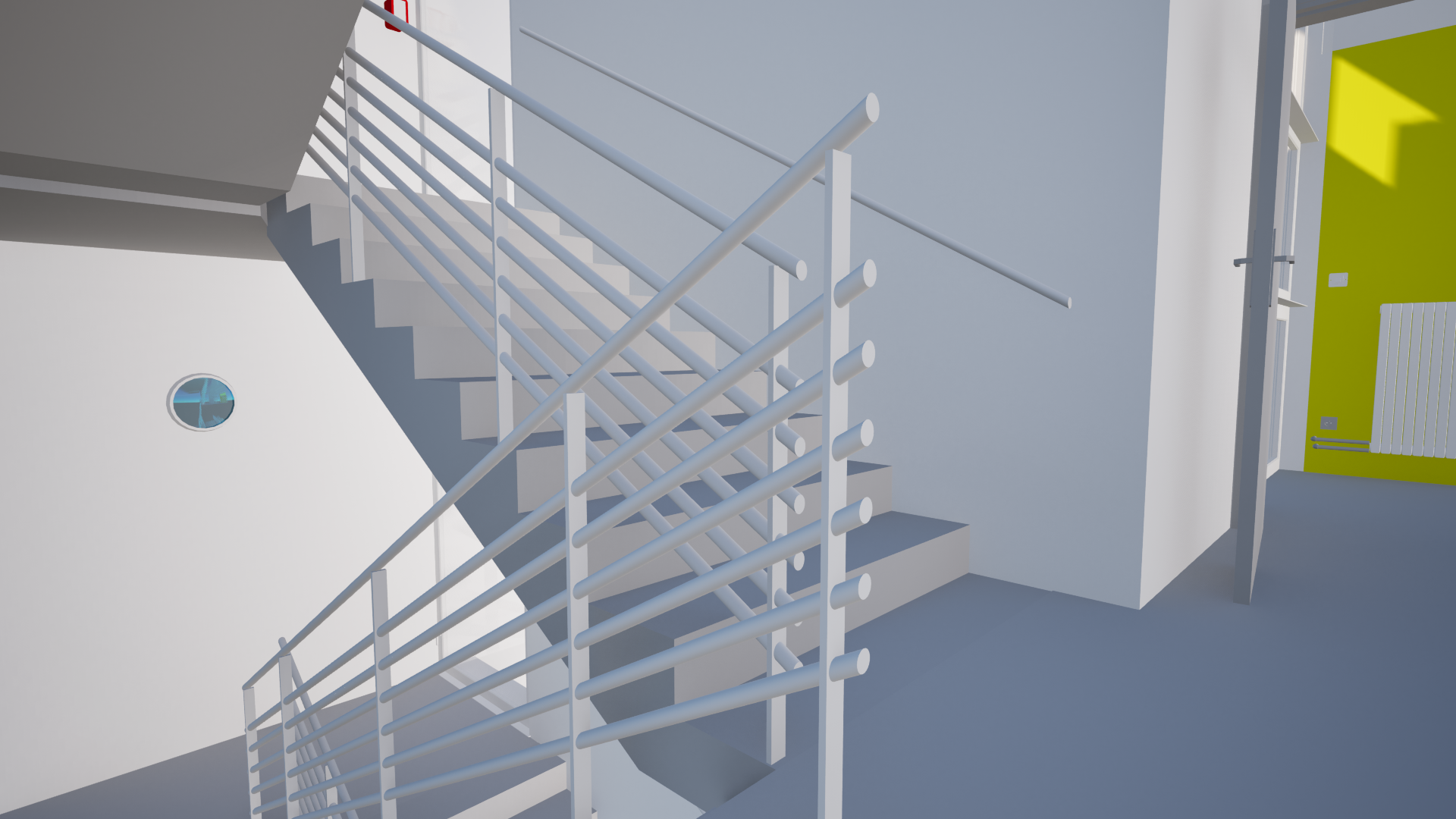}
        \caption{Structural objects: stairs, railings, doors, walls, floors.}\vspace{3pt}
        \label{fig:model_structure}
    \end{subfigure}

    \begin{subfigure}[b]{0.225\textwidth}
        \centering
        \includegraphics[width=\textwidth]{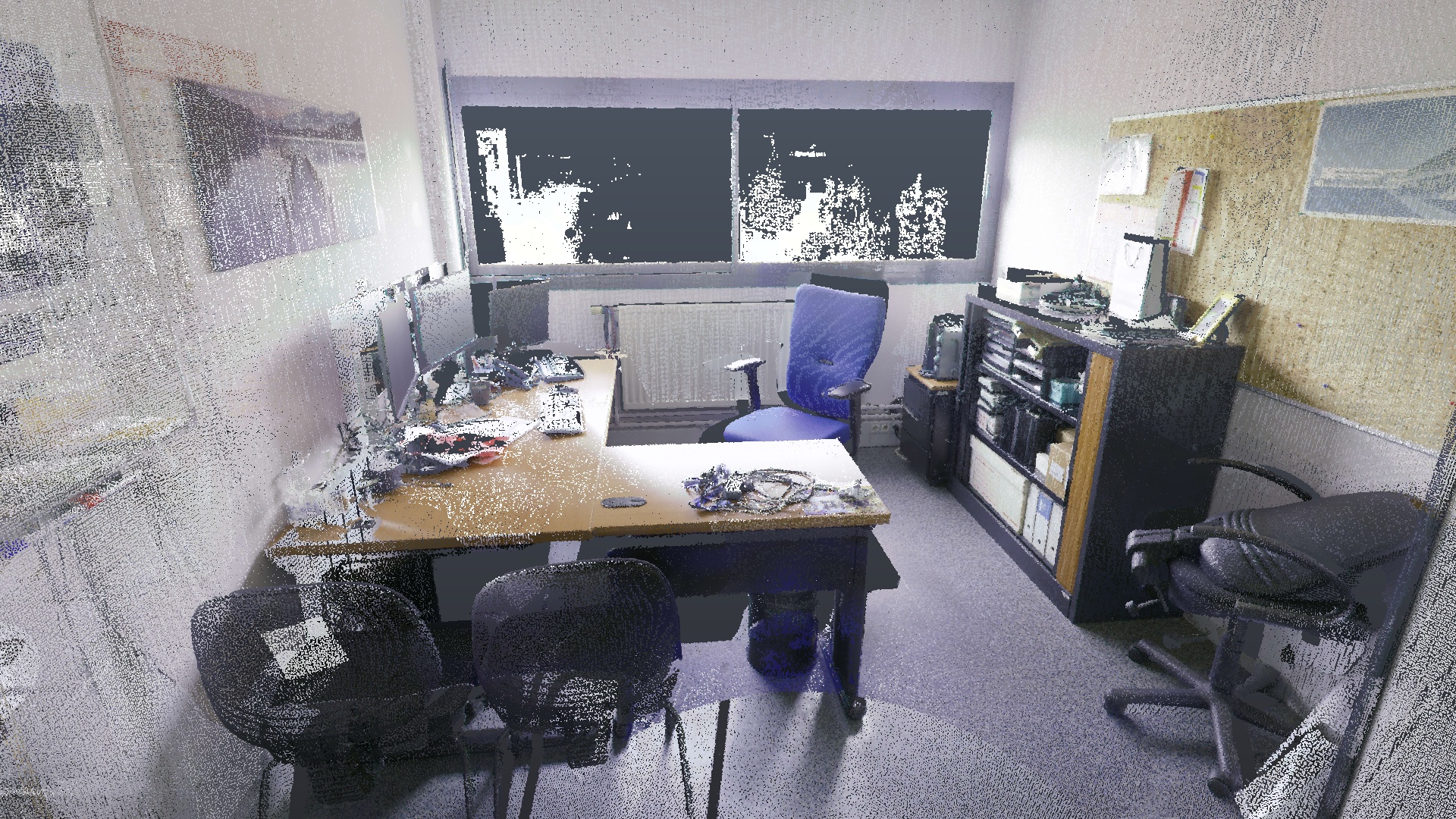}
        \caption{3D point cloud of a room}\vspace{3pt}
        \label{fig:room_pointcloud}
    \end{subfigure}%
    \begin{subfigure}[b]{0.225\textwidth}
        \centering
        \includegraphics[width=\textwidth]{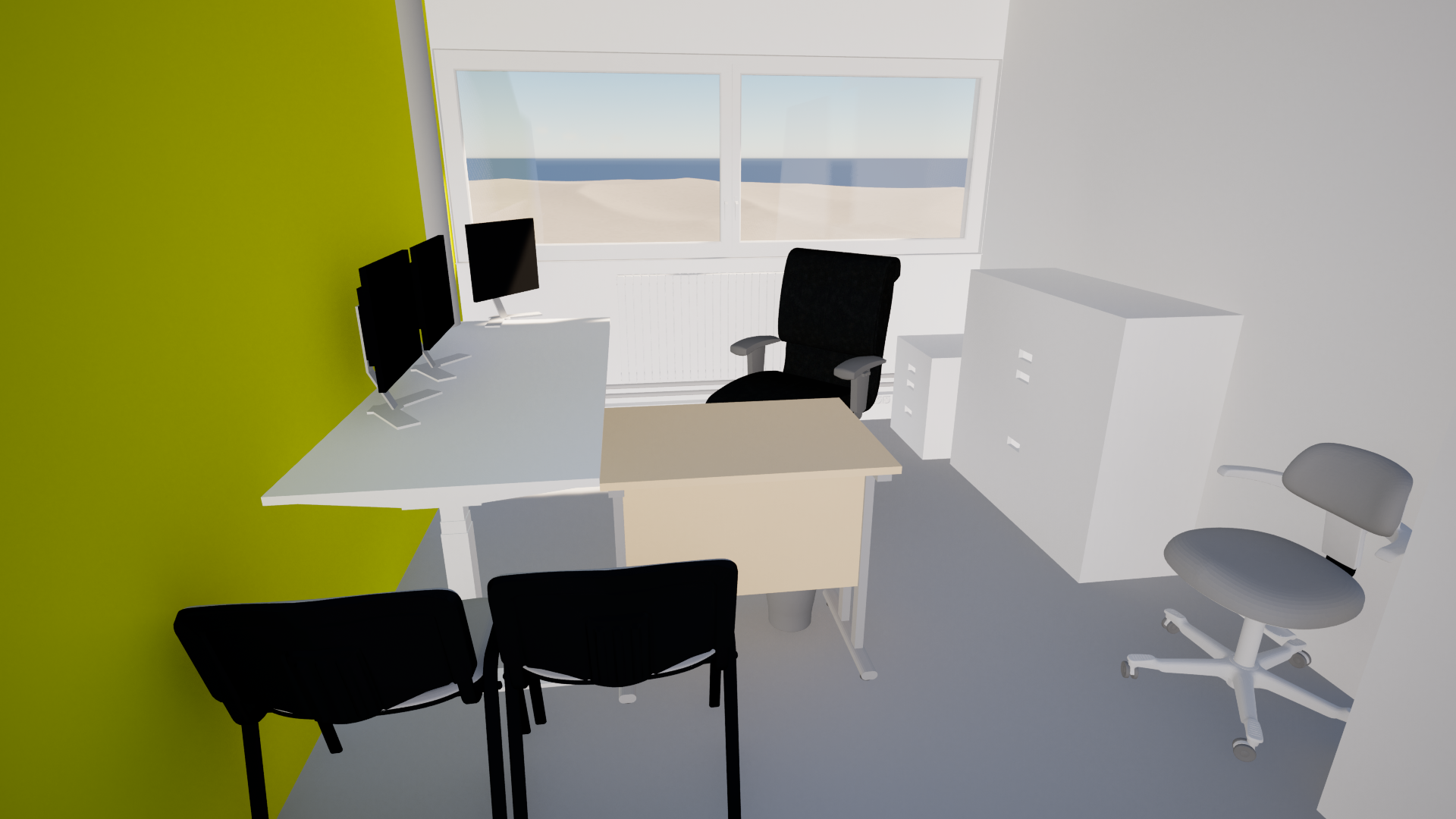}
        \caption{3D model of a room}\vspace{3pt}
        \label{fig:room_model}
    \end{subfigure}
    
    \begin{subfigure}[b]{0.45\textwidth}
        \centering
        \includegraphics[width=\textwidth]{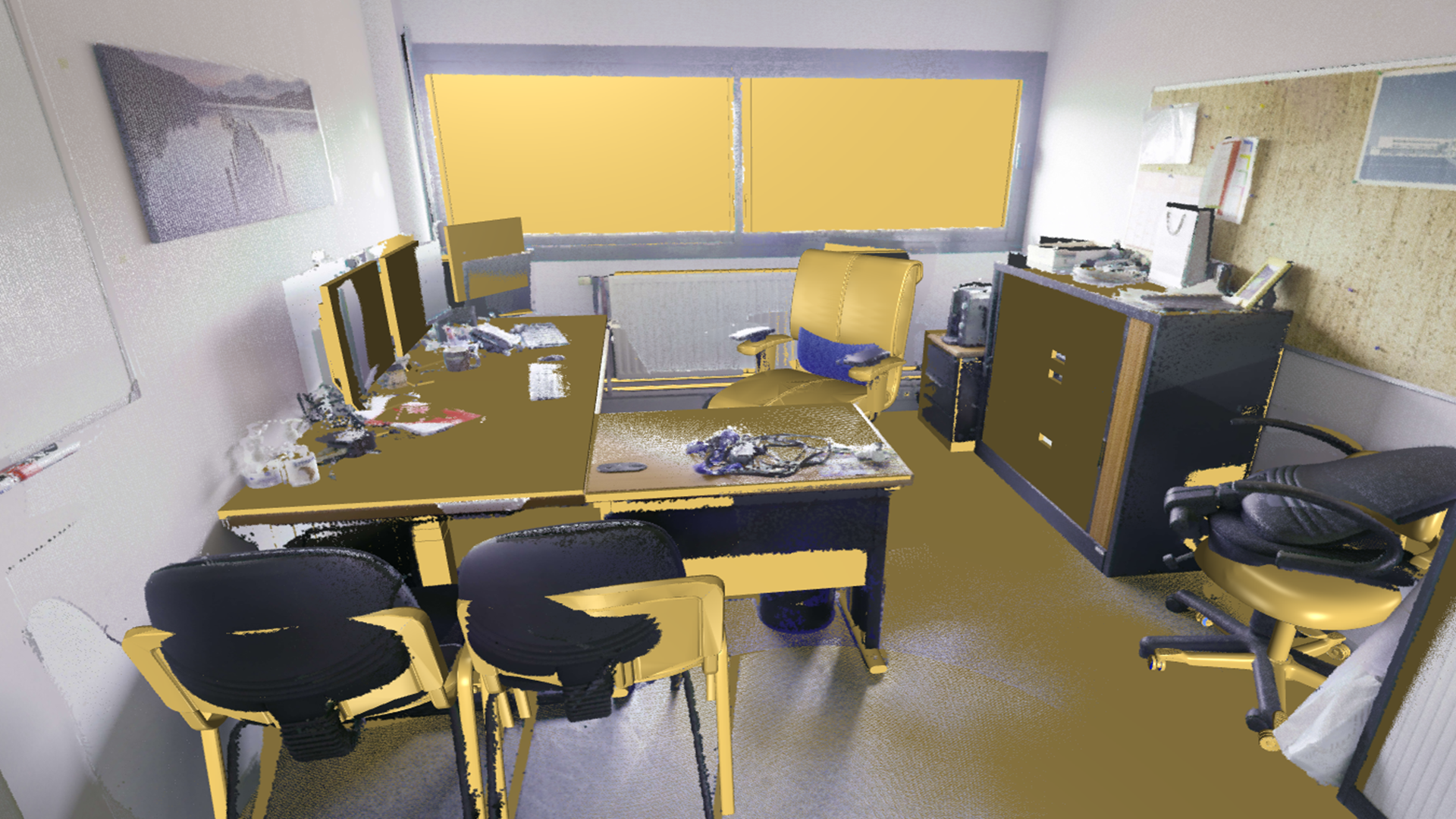}
        \caption{Overlay of clouds and objects}\vspace{3pt}
        \label{fig:room_overlay}
    \end{subfigure}

    \caption{View of a test area room. The generic 3D models are close, but not perfect matches for the actual scans.}
    \label{fig:model_pointcloud_mismatch}
\end{figure}

%% file: Table/Dataset_available.tex
\begin{table*}[t]

    \caption{Characteristics of the three variants of the 3DSES dataset.}
    \label{tab:dataset_details}

    \begin{tabularx}{\textwidth}{Ycccccc}
        \toprule
         \textbf{Variant} & \textbf{Scans} & \textbf{Points} & \textbf{Ground Truth} & \textbf{Pseudo-labels} &  \textbf{Features} & \textbf{Classes} \\
         \midrule
        Gold \emoji{1st-place-medal} & 10 & \num{65214193} & \cmark & \cmark & RGB \& I & 18 \\
        Silver \emoji{2nd-place-medal} & 30 & \num{216181580} & \cmark & \cmark & RGB \& I & 12\\ %
        Bronze \emoji{3rd-place-medal} & 42 & \num{413486927} & \xmark & \cmark & RGB \& I & 12 \\
        \bottomrule
    \end{tabularx}
\end{table*}

%% file: Histograms/Gold_repartion_classes.tex
\definecolor{bronze}{rgb}{0.8, 0.5, 0.2}
\definecolor{silver}{rgb}{0.75, 0.75, 0.75}
\definecolor{goldenyellow}{rgb}{1.0, 0.84, 0.0}

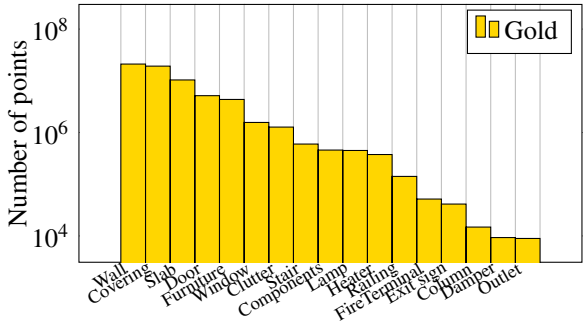
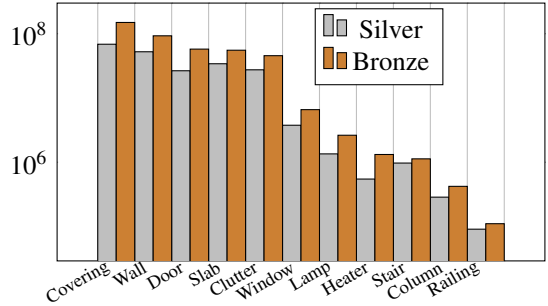
\begin{figure*}[t]
    \centering
    \hspace{-1.2cm}
    \begin{subfigure}[t]{0.55\textwidth}
        \begin{tikzpicture}
            \begin{axis}[
                ymode=log,
                ybar interval,
                xlabel style={yshift=0pt},
                ylabel=Number of points,
                ylabel style={yshift=-0.25cm},
                xticklabels= {Wall, Covering, Slab, Door, Furniture, Window, Clutter, Stair, Components, Lamp, Heater, Railing, FireTerminal, Exit sign, Column, Damper, Outlet, Switch},
                xtick={0,...,17},
                tickwidth = 0pt,
                height=5cm,
                width=8.2cm,
                ymax=3e8,
                xticklabel style={rotate=30, anchor=east,font=\scriptsize} %
            ]
            \addplot [fill=goldenyellow] coordinates{(0,21080009) (1,19230959) (2,10388198) (3,5157449) (4,4361248) (5,1565795) (6,1273661) (7,598135) (8,457574) (9,449462) (10,375239) (11,141609) (12,51663) (13,41334) (14,14835) (15,9294) (16,8936) (17,8793)};
            \legend{Gold}
            \end{axis}
        \end{tikzpicture}
        \caption{Real labels distribution (Gold).}
        \label{fig:histogram_gold}
    \end{subfigure}%
    \begin{subfigure}[t]{0.45\textwidth}
        \centering
        \begin{tikzpicture}
            \begin{axis}[
                ymode=log,
                ybar interval,
                xlabel style={yshift=0pt},
                ylabel style={yshift=0pt},
                xticklabels={Covering, Wall, Door, Slab, Clutter, Window, Lamp, Heater, Stair, Column, Railing, Exit sign },
                xtick={0,...,11},
                height=5cm,
                width=8cm,
                tickwidth= 0pt,
                ymax=3e8,
                legend style={at={(0.66,0.98)},
 anchor=north},
                xticklabel style={rotate=30, anchor=east, font=\scriptsize} %
            ]

            ]
            
            \addplot [fill= silver] coordinates{(0,68655724)(1,52391490)(2,26522716)(3,34097234)(4,27439944)(5,3765049)(6,1347990)(7,546894)(8,970494)(9,285928)(10,90713)(11,67404)}; %

            \addplot  [fill=bronze] coordinates{(0,149817740) (1,92833002) (2,57632155) (3,55392912) (4,45527881) (5,6583113) (6,2630245) (7,1319162) (8,1128388) (9,421235) (10,110387) (11,90707)};

            \legend{Silver, Bronze}
            \end{axis}
        \end{tikzpicture}
        \caption{Pseudo-labels (Silver/Bronze).}
        \label{fig:histogram_silverbronze}
    \end{subfigure}
    \caption{Distribution of the real and pseudo labels in the variants of the 3DSES dataset.}
    \label{fig:class_histograms}
\end{figure*}

%% file: Table/Evaluation_Pseudo_Label.tex
\begin{table*}[t]
\begin{center}
    \caption{Evaluation of the accuracy of the pseudo-labels obtained using our alignement algorithm on 3DSES. %
    Intersection over Union (IoU) per class, mean IoU (mIoU), overall accuracy (OA) and average accuracy (AA).}

        \label{tab:matching_performances}
        \newcommand{\tableclasslabel}[1]{\rotatebox[origin=c]{30}{\scriptsize #1}}
        \setlength\tabcolsep{2pt}
       
        \scriptsize
        \begin{tabularx}{\textwidth}{lYYYYYYYYYYYYYYYYYYccc}
        
        \toprule
        \tableclasslabel{\textbf{Variant}}&
        \tableclasslabel{Column} &
        \tableclasslabel{Components} &
        \tableclasslabel{Covering} &
        \tableclasslabel{Damper} &
        \tableclasslabel{Door} &
        \tableclasslabel{Exit sign} &
        \tableclasslabel{Fire terminal} &
        \tableclasslabel{Furniture} &
        \tableclasslabel{Heater} &
        \tableclasslabel{Lamp} &
        \tableclasslabel{Outlet} &
        \tableclasslabel{Railing} &
        \tableclasslabel{Slab} &
        \tableclasslabel{Stair} &
        \tableclasslabel{Switch} &
        \tableclasslabel{Wall} &
        \tableclasslabel{Window} &
        \tableclasslabel{Clutter} &
        \footnotesize \textbf{OA} &
        \footnotesize \textbf{AA} &
        \footnotesize \textbf{mIoU}\\
        \midrule
        \textbf{Gold}&
        21.00 & 80.96 & 95.95 & 77.29 & 91.95 & 73.16 & 86.57 & 79.48 & 91.08 & 66.71 & 37.59 & 58.52 & 95.05 & 59.07 & 45.66 & 93.64 & 64.55 & 36.44 &
        \footnotesize 94.66 &
        \footnotesize 83.09 &
        \footnotesize 69.70 \\
        
        \bottomrule
        \textbf{Silver}&
        25.02 & \xmark & 97.99 & \xmark & 93.97 & 72.27 & \xmark & \xmark & 82.22 & 73.88 & \xmark & 58.52 & 96.20 & 59.07 & \xmark & 91.52 & 56.67 & 88.88 & 
        \footnotesize 96.37 &
        \footnotesize 83.40 &
        \footnotesize 74.68\\
        
        \bottomrule
        \end{tabularx}

\end{center}
\end{table*}

%% file: sec/4_Experiments.tex
    To assess the difficulty of 3DSES, we evaluate initial baselines for the three variants: Gold, Silver and Bronze. We opt for PointNeXt \cite{NEURIPS2022_9318763d} and Swin3D \cite{yang_swin3d_2023}, since they are some of the highest performing models for semantic segmentation on S3DIS \cite{armeni_cvpr16}, and their code is available. %
    We compare PointNeXt-S (\num{800000} parameters) to Swin3D-L (68M parameters).

    Note that these models both perform voxelization and therefore do not benefit from the extremely high point density of 3DSES.
    In particular, PointNeXt is not designed to process dense point clouds in optimum time ($\approx4$ hours per scan). To reduce inference times, we subsample our test point clouds to \SI{1}{\centi\meter}.
    We expect that future models evaluated on 3DSES will better take into account the fine resolution of indoor TLS scans.

    \paragraph{Hyperparameters} We train Swin3D-L with AdamW, a cosine learning rate for 100 epochs, a batch size of 6, and an inverse class frequency weighted cross-entropy to deal with class imbalance. PointNext-S is trained with the original S3DIS hyperparameters: epochs $=100$, batch size $=32$, AdamW optimizer, a CosineScheduler and a non-reweighted CrossEntropyLoss. We only tune the learning rate to $l_r=0.05$ (instead of 0.01 in original setup). Following standard practices \cite{wang_o-cnn_2017,wu_point_2022,yang_swin3d_2023},
     we use test-time augmentation and aggregate segmentation predictions with a majority vote over 12 rotations.
     Models are trained on an NVIDIA RTX A6000

    \input{Table/Bencharking_GoldDataset}

    \paragraph{Results on 3DSES Gold} We train both Swin3D-L and PointNeXt-S models on 3DSES Gold: one on the real labels and the other on the pseudo-labels. All models are evaluated on the ground truth over the test area. Results are reported in \cref{tab:gold_benchmark_intensity}. We observe that 3DSES is a challenging dataset: mean IoU is heavily penalized by performance on small objects. Classes comprised of small objects with few points ($<10^5$ points) are difficult to learn and the model either never predicts them, or makes significant errors.
    Note that despite its high intraclass variance, ``Clutter'' is mostly well segmented with a $>50\%$ IoU, showing that the model is able to automatically identify most irrelevant objects from the point clouds.
    Interestingly, the results also show that Swin3D only slightly underperforms when trained on the pseudo-labels, with a $1.2$\% decrease in mIoU (47.8\% vs. 49.0\%) compared to the model trained on the real labels. Segmentation errors when using pseudo-labels are concentrated on classes for which the alignment procedure showed weaknesses, such as ``Stair'' and ``Railing''. %
    This demonstrates the potential of using CAD models to automatically label point clouds, as way of circumventing the lack of annotated datasets for specialized settings (\ie factories, schools or administrative buildings\dots). %
    PointNext struggles with 3DSES and achieves low mIoU scores.
    However, the same trends hold with better segmentation of structural elements and underperformance on minority classes.

    \paragraph{Results on Silver/Bronze} We report in \cref{tab:silver_benchmark_intensity} the segmentation metrics on the 3DSES test set when training Swin3D and PointNext on Silver, both with pseudo and real labels, and on Bronze with pseudo labels. We observe that metrics are consistently higher for all 12 classes on Silver with real label compared to training the Gold subset. This is expected, since the Silver classification is simpler and removes small objects that were heavily penalized. Yet, the larger training set (Silver is $3\times$ as large as Gold) benefits the segmentation, with higher scores on the ``Lamp'', ``Window'' and ``Clutter'' classes that exhibit strong diversity.
    Training with pseudo-labels on Silver results in a significant performance drop, correlated with the lower class alignment scores discussed in \cref{Matching_score}.
    Yet results on 3DSES Bronze show that the noise in the pseudo-labels can be alleviated by a larger dataset. Despite using raw point clouds and error-prone pseudo-labels, models trained on Bronze achieves similar (PointNeXt) or even better (Swin3D) segmentation accuracy than when trained on the clean Silver dataset.
    We assume that diversity partially compensates for label noise, allowing models to learn better invariances despite small errors in the labels. In addition, the raw point clouds are denser that the clean versions used in Silver and Bronze and might provide more geometrical information that is more costly to process, but also more discriminative.
    These observations show the tradeoffs of the three variants of 3DSES, from training on small high-quality data, to larger but noisier point clouds.

    \input{Table/BenchmarkingSilverBronze}

    \paragraph{Impact of Lidar intensity} As described in \cref{sec:related}, 3DSES is the only indoor TLS dataset that provides Lidar intensity.
    We included intensity as an additional feature in our models to evaluate its impact on semantic segmentation.
    As shown in \cref{tab:gold_benchmark_intensity} for Swin3D, we observe a $3.0$\% increase in mIoU when using intensity in addition to color on real labels. 
    Nonetheless, we observe a decrease for Swin3D on pseudo-labels ($2.6$\%). However, the drop is not consistent on all classes, \eg few classes obtain better IoU.
    On the other hand, including the intensity for PointNeXt improves mIoU by $15\%$.
    This shows that intensity helps generalization of smaller models.
    In \cref{tab:silver_benchmark_intensity}, intensity helps Swin3D and PointNeXt in most cases. %
    In comparison, Swin3D trained on Silver variant with pseudo-labels and intensity obtains \emph{better} scores ($+12.7$\% IoU) than without intensity.
    Overall, the preliminary results could indicate that Lidar intensity can indeed be discriminative for some classes, especially for larger datasets.
    Further experiments are required to validate these observations.%

%% file: Table/Bencharking_GoldDataset.tex
\begin{table*}[t]
\begin{center}

\caption{Segmentation metrics on the test set for 3DSES Gold, either with real or pseudo labels (and intensity features or not). Intersection over union (IoU) per class, mean IoU (mIoU), overall accuracy (OA), average accuracy (AA).}
\label{tab:gold_benchmark_intensity}

\newcommand{\tableclasslabel}[1]{\rotatebox[origin=c]{30}{\scriptsize #1}}
\setlength\tabcolsep{2pt}

\scriptsize
\begin{tabularx}{\textwidth}{lXXXXXXXXXXXXXXXXXXXXccc}
\toprule
&
\tableclasslabel{\textbf{Real labels}} &
\tableclasslabel{\textbf{Intensity}} &
\tableclasslabel{Column} &
\tableclasslabel{Components} &
\tableclasslabel{Covering} &
\tableclasslabel{Damper} &
\tableclasslabel{Door} &
\tableclasslabel{Exit sign} &
\tableclasslabel{Fire terminal} &
\tableclasslabel{Furniture} &
\tableclasslabel{Heater} &
\tableclasslabel{Lamp} &
\tableclasslabel{Outlet} &
\tableclasslabel{Railing} &
\tableclasslabel{Slab} &
\tableclasslabel{Stair} &
\tableclasslabel{Switch} &
\tableclasslabel{Wall} &
\tableclasslabel{Window} &
\tableclasslabel{Clutter} &
\footnotesize \textbf{OA} &
\footnotesize \textbf{AA} &
\footnotesize \textbf{IoU}\\
\midrule
\multirow{4}{*}{\rotatebox[origin=c]{90}{Swin3D}} &
\footnotesize\cmark &
\footnotesize\xmark &
0.00 &
31.16 &
90.12 &
14.63 &
75.95 &
12.19 &
56.67 &
71.57 &
76.18 &
26.76 &
9.53 &
71.75 &
87.63 &
70.59 &
0.00 &
88.40 &
47.26 &
52.03 &
\footnotesize 89.74 &
\footnotesize 78.30 &
\footnotesize 49.02 \\

&
\footnotesize\cmark &
\footnotesize\cmark &
0.00  &
49.76 &
94.62 &
18.23 &
81.87 &
27.37 &
67.10  &
73.13 &
83.61 &
47.73 &
0.00  &
57.31 &
85.29 &
56.67 &
0.00  &
89.68 &
53.54 &
50.46 &
\footnotesize 91.64 &
\footnotesize 74.45 &
\footnotesize 52.02 \\

&
\footnotesize\xmark &
\footnotesize\xmark &
17.52 &
34.81 &
88.90 &
31.71 &
75.84 &
16.31 &
48.28 &
68.87 &
71.04 &
24.50 &
12.85 &
45.53 &
86.84 &
58.64 &
0.93 &
87.09 &
50.59 &
40.31 &
\footnotesize 88.54 &
\footnotesize 76.80 &
\footnotesize 47.81 \\

&
\footnotesize\xmark &
\footnotesize\cmark &
30.06 &
51.07 &
93.29 &
63.98 &
54.16 &
0.00 &
21.36 &
51.32 &
66.14 &
41.09 &
6.33 &
50.31 &
79.04 &
40.46 &
0.00 &
83.92 &
48.96 &
31.98 &
\footnotesize 86.48 &
\footnotesize 74.10 &
\footnotesize 45.19 \\

\midrule
\multirow{4}{*}{\rotatebox[origin=c]{90}{PointNeXt-S}} &
\footnotesize\cmark &
\footnotesize\xmark &
0.00 &
0.00 &
96.27 &
0.00 &
35.43 &
0.00 &
0.00 &
32.84 &
0.00 &
69.12 &
0.00 &
0.00 &
90.87 &
60.40 &
0.00 &
74.58 &
38.05 &
24.80 &
\footnotesize 82.58 &
\footnotesize 35.04 &
\footnotesize 29.02 \\

&
\footnotesize\cmark &
\footnotesize\cmark &
0.00 &
56.16 &
96.73 &
0.00 &
65.80 &
0.00 &
0.00 &
52.57 &
26.59 &
72.78 &
0.00 &
60.75 &
94.28 &
85.93 &
0.00 &
86.76 &
59.78 &
39.47 &
\footnotesize 91.19 &
\footnotesize 49.25 &
\footnotesize 44.31 \\

&
\footnotesize\xmark &
\footnotesize\xmark &
0.00 &
0.01 &
96.01 &
0.00 &
37.57 &
0.00 &
0.00 &
45.11 &
0.00 &
39.76 &
0.00 &
0.00 &
89.73 &
60.33 &
0.00 &
77.57 &
1.18 &
20.33 &
\footnotesize 84.19 &
\footnotesize 30.48 &
\footnotesize 25.98 \\

&
\footnotesize\xmark &
\footnotesize\cmark &
0.00 &
50.10 &
96.68 &
0.00 &
67.86 &
0.00 &
0.00 &
49.83 &
43.32 &
65.51 &
0.00 &
7.51 &
93.79 &
81.23 &
0.00 &
86.27 &
55.81 &
21.35 &
\footnotesize 90.08 &
\footnotesize 44.86 &
\footnotesize 39.96 \\
\bottomrule
\end{tabularx}
\end{center}
\end{table*}

%% file: Table/BenchmarkingSilverBronze.tex
\begin{table*}[t]
\begin{center}

\caption{Segmentation metrics on the test set for 3DSES Silver and Bronze, either with real or pseudo labels (and intensity features or not). Intersection over union (IoU) per class, mean IoU (mIoU), overall accuracy (OA), average accuracy (AA).}
\label{tab:silver_benchmark_intensity}
\newcommand{\tableclasslabel}[1]{\rotatebox[origin=c]{0}{\scriptsize #1}}
\setlength\tabcolsep{2pt}

\scriptsize
\begin{tabularx}{\textwidth}{ccYYYYYYYYYYYYYYccc}

\toprule
& &
\tableclasslabel{\textbf{Labels}} &
\tableclasslabel{\hspace{-2pt}\textbf{Intensity}}&
\tableclasslabel{Column} &
\tableclasslabel{Covering} &
\tableclasslabel{Door} &
\tableclasslabel{Exit sign} &
\tableclasslabel{Heater} &
\tableclasslabel{Lamp} &
\tableclasslabel{Railing} &
\tableclasslabel{Slab} &
\tableclasslabel{Stair} &
\tableclasslabel{Wall} &
\tableclasslabel{Window} &
\tableclasslabel{Clutter} &
\footnotesize \textbf{OA} &
\footnotesize \textbf{AA} &
\footnotesize \textbf{IoU}\\
\midrule

\multirow{6}{*}{\rotatebox[origin=c]{90}{Swin3D}} &

    \multirow{4}{*}{\rotatebox[origin=c]{90}{Silver}} &
    	\footnotesize\cmark & \footnotesize\xmark & 0.00 & 89.07 & 76.40 & 9.93 & 74.69 & 32.24 & 46.22 & 86.40 & 67.75 & 89.24 & 54.62 & 90.42 & \footnotesize 91.69 & \footnotesize 84.84 & \footnotesize 59.75\\
        & & \footnotesize\cmark & \footnotesize\cmark &
        5.40 & 94.35 & 83.06 & 9.30 & 75.27 & 44.04 & 37.63 &  84.08 & 38.69 & 85.34 & 54.99 & 72.83 &\footnotesize 90.47 & \footnotesize 83.39 & \footnotesize 57.08\\
        & &\footnotesize\xmark & \footnotesize\xmark & 25.47 & 88.50 & 61.62 & 12.96 & 59.24 & 30.79 & 35.94 & 77.55 & 36.22 & 87.61 & 48.76 & 71.15 & \footnotesize 87.49 & \footnotesize 88.44 & \footnotesize 52.98\\
        & &\footnotesize\xmark & \footnotesize\cmark &
        52.31 & 95.82 & 89.01 & 11.79 & 65.29 & 55.28 & 64.17 & 82.06 & 34.32 & 92.44 & 54.00 & 91.92 &\footnotesize 93.46 & \footnotesize 89.44 & \footnotesize 65.70\\
    
    \cmidrule(l){2-19}
    &
    \multirow{2}{*}{\rotatebox[origin=c]{90}{Bronze}} &
    	\footnotesize\xmark & \footnotesize\xmark &
    	51.76 & 95.90 & 89.37 & 12.45 & 65.80 & 52.25 & 82.14 & 86.80 & 43.15 & 93.33 & 60.53 & 93.59 & \footnotesize 94.59 & \footnotesize 93.67 & \footnotesize 68.92\\
        & &\footnotesize\xmark & \footnotesize\cmark &
    	59.68 & 95.97 & 88.10 & 41.80 & 71.59 & 55.59 & 77.20 & 85.81 & 41.40 & 93.00 & 60.89 & 94.52 & \footnotesize 94.51 & \footnotesize 94.37 & \footnotesize 72.13\\
	\midrule
\multirow{6}{*}{\rotatebox[origin=c]{90}{PointNeXt-S}} &

    \multirow{4}{*}{\rotatebox[origin=c]{90}{Silver}}&
    \footnotesize\cmark & \footnotesize\xmark & 0.00 & 96.77 & 67.11 & 0.00 & 16.45 & 69.95 & 61.75 & 94.88 & 83.87 & 89.26 & 62.54 & 80.25 & \footnotesize 93.30 & \footnotesize 66.27 & \footnotesize 60.24\\
    & & \footnotesize\cmark & \footnotesize\cmark &
    0.00 & 97.07 & 76.66 & 0.00 & 38.73 & 78.11 & 65.26 & 94.85 & 86.97 & 90.84 & 67.08 & 84.35 & \footnotesize 94.63 & \footnotesize 70.59 & \footnotesize 64.99\\

    & &\footnotesize\xmark & \footnotesize\xmark & 
    0.00 & 96.53 & 73.07 & 0.00 & 20.33 & 66.71 & 2.79 & 93.50 & 76.90 & 90.32 & 40.60 & 71.12 & \footnotesize 92.68 & \footnotesize 57.60 & \footnotesize 52.66\\
    & &\footnotesize\xmark & \footnotesize\cmark &
    58.44 & 96.55 & 69.81 & 0.00 & 33.96 & 67.00 & 38.90 & 93.86 & 83.48 & 88.12 & 51.25 & 73.60 & \footnotesize 92.58 & \footnotesize 71.19 & \footnotesize 62.91\\

    \cmidrule(l){2-19}
    &
        \multirow{2}{*}{\rotatebox[origin=c]{90}{Bronze}}&
        \footnotesize\xmark & \footnotesize\xmark &
        11.21 & 95.68 & 85.16 & 0.00 & 69.18 & 66.19 & 15.97 & 93.53 & 80.09 & 92.62 & 49.09 & 82.86 & \footnotesize 94.57 & \footnotesize 66.47 & \footnotesize 61.79\\
        &&\footnotesize\xmark & \footnotesize\cmark &
        56.45 & 96.44 & 81.39 & 0.00 & 79.71 & 77.40 & 42.25 & 93.35 & 78.33 & 91.57 & 56.47 & 80.94 & \footnotesize 94.47 & \footnotesize 77.06 & \footnotesize 69.53\\
    \bottomrule

    \end{tabularx}

\end{center}
\end{table*}

%% file: sec/5_Conclusion.tex
We introduced 3DSES, a new dataset for semantic segmentation of dense indoor Lidar point cloud. 3DSES fills the need for indoor TLS datasets designed for building survey and modeling. It contains a unique combination of point cloud labels for semantic segmentation, a georeferenced 3D CAD model with BIM oriented objects and Lidar intensity, a radiometric feature not provided in existing datasets. We demonstrate that using 3D CAD models to automatically annotate point clouds is a time-efficient strategy that produces pseudo-labels with 95\% accuracy compared to a manual ground truth. Moreover, we show that training on pseudo-labels achieves similar performance to training on real ones on 3DSES.
We show that segmentation accuracy can benefit from Lidar intensity in indoor settings, despite radiometry being often ignored in previous works.
Segmentation results demonstrate that 3DSES is a challenging new dataset, especially for BIM-oriented classes, \eg small building components such as electrical terminals and safety systems. We hope this new dataset will stimulate research on indoor point clouds processing and motivate the community to investigate auto-modeling tasks in scan-to-BIM.

%% file: sec/X_suppl.tex
\section{Dataset structure}

\input{Image/Vue_dessus}

    All the information presented in this section is reproduced in the ``README'' file of the 3DSES dataset archive.
    The dataset is hosted on \href{https://zenodo.org}{Zenodo} for public release under the Creative Commons CC-BY-SA 4.0 license at the following URL: %
    \href{https://zenodo.org/records/13323342?preview=1&token=eyJhbGciOiJIUzUxMiJ9.eyJpZCI6IjY4YzNhZTY5LTRlY2QtNDM3ZS1hYTYxLTgyYzA2ZWU3Y2U5YyIsImRhdGEiOnt9LCJyYW5kb20iOiI0ODQ1Nzk4OGQ4ZTc4NzVjYTc0ZWZkODFmZTIzNGIwMiJ9.RAfbzBH2rM82BUFhmQKncbzH_cxXcLcpu8D_aG0zVC97VVqob6jreCR55Mj6SJHGTEOivgE4ZY5ys2x3VgkP2g}{\texttt{\censor{https://zenodo.org/records/13323342}}}.
    The \texttt{zip} archive contains three folders, one for each variant (see \cref{fig:topview_3dses}):
    \begin{itemize}
        \item '\texttt{Gold/}': contains Gold point clouds,
        \item '\texttt{Silver/}': contains Silver point clouds,
        \item '\texttt{Bronze/}': contains all raw point clouds\footnote{Directly exported from Register360.} for the Bronze version.
    \end{itemize}
    We provide our three variants of 3DSES: Gold, Silver and Bronze. We use the NumPy \cite{harris2020array} \texttt{.npy} format to store the point clouds. Points clouds are organized per scan, identified by \texttt{SXXX}, where \texttt{XXX} is a three-digits integer. Three test points clouds are currently kept private: scans \texttt{S170}, \texttt{S171} and \texttt{S180}. These point clouds are kept hidden for use as evaluation on a future \href{https://www.codabench.org/}{Codabench} competition. Point clouds can be opened with NumPy  using Python, \eg with \texttt{numpy.load}.
    
    For Gold and Silver versions, the \texttt{.npy} files contain 9 columns, where each row describe one point in the scan. Column signification is detailed in \cref{tab:dataset_features}.
    
    Labels, for Gold version, are in the range of $\llbracket 0, 17 \rrbracket$. For the Silver versions, labels are in the range $\llbracket 0, 11 \rrbracket$ instead since the classification uses only 12 classes.
    
    Since the Bronze variant of 3DSES only contains pseudo-labels, column \#7 contains instead the pseudo label for these scans, and column \#8 is dropped.

    \input{Table/column_signification}

    \paragraph{Preprocessing}
    We provide unnormalized color information, \ie RGB values are comprised in $[0-255]$. Regarding the $(xyz)$ coordinates, we apply a translation from the initial georeferenced point cloud to obtained centered and smaller coordinates that are more convenient for use in deep learning models.\\
    
\begin{figure}[t]
        \centering
        \includegraphics[width=0.5\textwidth]{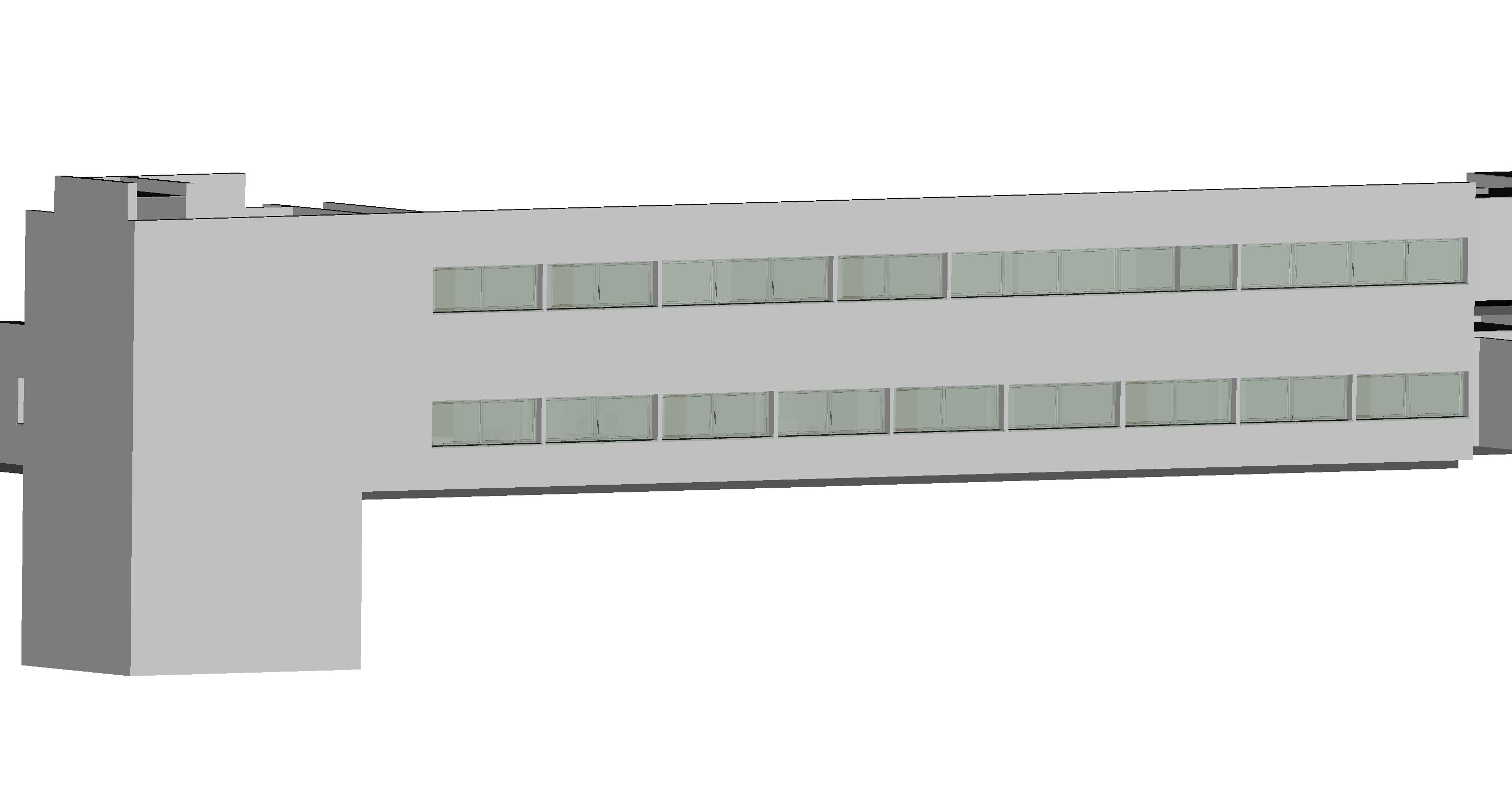}%
        \caption{Screenshot of 3D model of 3DSES inside CloudCompare viewer.}
        \label{fig:cloudcompare_capture}%
    \end{figure}

    \paragraph{CAD model} The CAD model is currently distributed as an \texttt{.obj} file and will be made available in an open format supporting the IFC standard for public release. This CAD model, can be visualize with a 3D data processing software (such as CloudCompare \cite{girardeau-montaut_detection_2006}, see \cref{fig:cloudcompare_capture}).

\section{Classification and label signification}\label{Suppmaterial_classification_details}

        \paragraph{Taxonomy}
        3DSES uses a BIM-oriented class taxonomy, that focuses on modeling both the structure of a building and its functional equipment.
        Classes composed of structural elements (\eg covering, slab, clutter) are the easiest to understand, since they are the technical terms for the common parts of a building: walls, floors, ceilings, \etc However, 3DSES also includes domain-specific classes that regroup many different types of objects. We have grouped such utilities by domain according to their purpose, \eg fire suppression,  heating, electrical systems, lighting, \etc. We detail in \cref{tab:classes} the definition of every class in the dataset. In addition, \cref{fire_terminal,lamp,components,other} show some examples of standard 3D models for different types of objects.

\input{Table/Gold_taxonomy}

\input{Suppmaterial/Image_maquette}

\input{Suppmaterial/Image_furniture}

        \paragraph{Simplification for the Silver and Bronze variants}
            The 3DSES Silver variant uses a less detailed classification than 3DSES\-Gold. In particular, it focuses more on structural elements. To obtain this simplified classification, we followed three principles:
            \begin{enumerate}
                \item remove all small individual objects,
                \item remove all objects that do not have a well defined 3D CAD model,
                \item remove all objects that are not widely represented in the point clouds.
            \end{enumerate}
            
            This resulted in the following simplifications applied the classification:
            \begin{itemize}
                \item we merge all objects from classes ``Outlet'' and ``Switch'' into the class ``Wall'',
                \item objects from the ``Damper'' class are either merged into ``Covering'' or the ``Clutter'' class, depending on their distance to the closest points from these classes (smoke detectors are usually mounted to the ceiling),
                \item objects from the ``Fire terminal'' class are either merged into ``Wall'' or the ``Clutter'' class, depending on their distance to the closest points from these classes (fire alarms are wall-mounted, extinguishers tend to stuck out),
                \item non-structural elements, \ie objects from the ``Component'' and ``Furniture'' classes, are reclassified as ``Clutter''.
            \end{itemize}
            
            Note that the only exception to the general simplification principles is the ``Exit sign'' class, since this class always represent the same object and is therefore accurately modeled in the CAD model, and is present in most scans due to safety regulations.
            For these reasons, we chose to keep the ``Exit sign'' class in the simplified classification of the Silver variant. %
            The final Silver/Bronze classification is summarized in \cref{tab:classes_simplified}.

\input{Table/Silver_taxonomy}

\section{Additional details on the pseudo-labeling alignment algorithm}

    As stated in the main paper, the pseudo-labels are generated using an alignment algorithm based on cloud-to-mesh distance computation. This distance uses the Metro algorithm \cite{cignoni_metro_1998}, as implementated in CloudCompare \cite{girardeau-montaut_detection_2006}.
    We give below some additional insights on this algorithm, its requirements and its accuracy.

\input{Algo/Algo}

    \subsection{Computational requirements} \label{Suppmaterial_computation_time}

    Operations on 3D point clouds tend to be computationally demanding, especially when density increases. To be useful, the pseudo-labeling strategy should be cost effective, with less dependency on human annotators, but also time effective.
    In practice, the bottleneck for pseudo-labeling based on the 3D model is the creation of the 3D model.
    Indeed, we present in Histogram \ref{histogramm_time} the processing times required to label the point clouds on the Gold version of 3DSES.
    All computations have been run on a consumer Intel(R) Xeon(R) CPU E5-2643 v4 @ \SI{3.40}{\giga\hertz}.

\input{Histograms/Computation}

    As can be expected, processing times to align the mesh to the point cloud vary depending on the complexity of the 3D shapes of the objects, and the size of the 3D point clouds. For example, pseudo-labeling of slab is significantly faster than on furniture. With \SI{1871.41}{\second} for ``Furniture'' against only \SI{1.36}{\second} for ``Slab''. It's $\approx$\SI{1376} times longer for our alignment methods to create ``Furniture'' pseudo-labels. In Histogram \ref{histogramm_time}, we observed that all complex object shapes, such as ``Exit sign'', ``Fire Terminal'', ``Furniture'' and ``Heater'' have computation time $>90 s$ (largely superior as other classes). Overall, it takes approximately 42 minutes to pseudo-label one scan. The complete process therefore takes about 7 hours to annotate all point clouds for the Gold version with 18 classes.
    Since the Silver and Bronze variants of 3DSES contain fewer classes, the processing time for the alignment algorithm is significantly lower. This is due in part to the absence of the ``Furniture'' in those datasets. In practice, Silver can be pseudo-labeled in less than 4 hours and half ($\approx9$ minutes per scan) and Bronze is pseudo-labeled in around 9 hours ($\approx13$ minutes per scan). The additional time required for Bronze version can be explained by the higher number of points in each scan compared to the Silver version.

    \subsection{Evaluation of the pseudo-labels} \label{Suppmaterial_confusion_matrix}

\input{Confusionmatrix/Confusion_matrix_Gold}

    In addition to the main metrics provided in the paper, we detail below the full confusion matrices between the pseudo-labels and the manually annotated ground truth for the Gold (\cref{tab:cm_gold}) and Silver (\cref{tab:cm_silver}) variants. Note that these two matrices are normalized by line.

    We observe in the confusion matrix that these points are classified as ``Wall''. Indeed, the alignment algorithms merge these small objects into the wall. Reducing the threshold for classifying points as ``Wall'' could alleviate this problem, but would in return generate more false positive ``Clutter'' points. Note also the relatively low score (\SI{60.6}{\%}) for ``Railing'', which can be explained by the mismatch between the 3D railings and the actual physical railings at \censor{ESGT}. The same observation holds partially for ``Window'', as the windows are modeled with a standard frame that does not perfectly match the actual windows.

\input{Confusionmatrix/Confusion_matrix_Silver}

%% file: Image/Vue_dessus.tex
\definecolor{bronze}{rgb}{0.8, 0.5, 0.2}
\definecolor{silver}{rgb}{0.75, 0.75, 0.75}
\definecolor{goldenyellow}{rgb}{1.0, 0.84, 0.0}

\begin{figure}[ht]
    \includegraphics[width=0.5\textwidth,trim=0 0 0 15,clip]{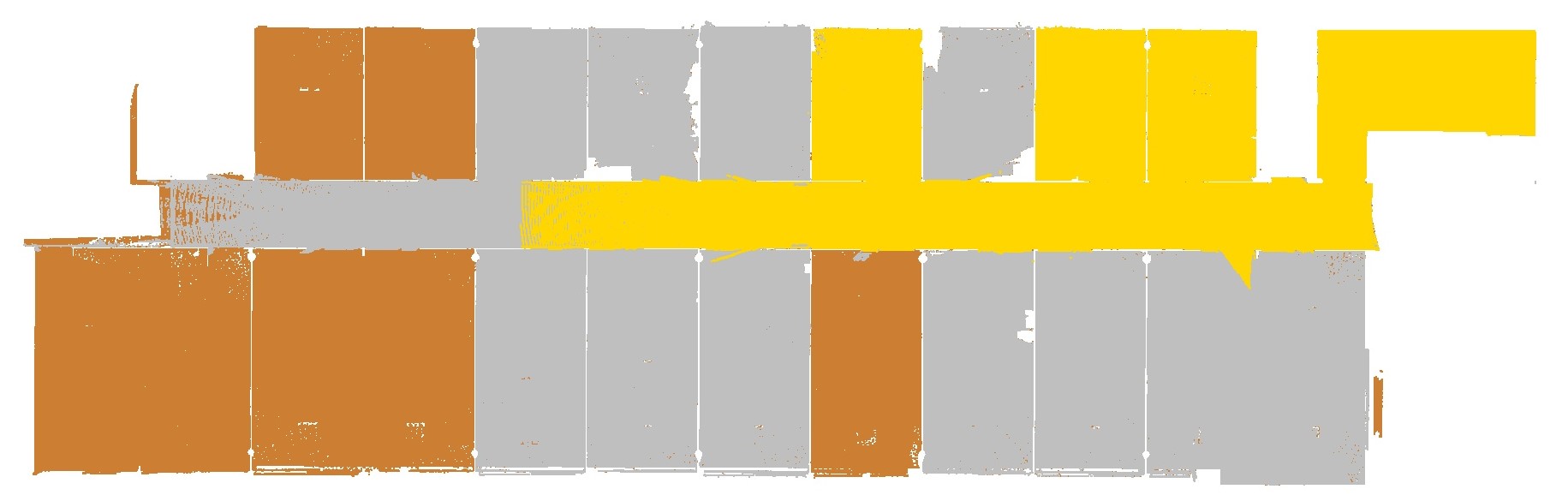}
    \caption{Top view of 3DSES and the split of points across the \textcolor{goldenyellow}{Gold}, \textcolor{silver}{Silver}, and \textcolor{bronze}{Bronze} variants. Note that these variants are inclusive,  \ie
    \textcolor{silver}{Silver} includes \textcolor{goldenyellow}{Gold} and \textcolor{bronze}{Bronze} includes \textcolor{silver}{Silver}.}
    \label{fig:topview_3dses}
\end{figure}

%% file: Table/column_signification.tex
\begin{table}[t]
\begin{center}
    \caption{Column signification in the point clouds files.}
    \begin{tabularx}{0.50\textwidth}{llX}
        \toprule
        \textbf{Index} & \textbf{Feature} & \textbf{Description}\\
        \midrule
        0   & $x$ & \multirow{3}{*}{
        \shortstack[l]{Point coordinates $(xyz)$ \\in an orthonormal basis \\with $z$ the height.}}\\
        1   & $y$ & \\
        2   & $z$ & \\
        \midrule
        3   & $r$ & \multirow{3}{*}{\shortstack[l]{Color in RGB format \\ encoded as uint8 $[0,255]$}}\\
        4   & $g$ & \\
        5   & $b$ & \\
        \midrule
        6   & Intensity & Lidar intensity encoded as float32 $[0,1]$\\
        \midrule
        7   & Real label & Manually annotated class in $\llbracket 0, n^\dagger \rrbracket$\\
        8   & Pseudo label & Automatically annotated class in $\llbracket 0, n^\dagger \rrbracket$\\
        \bottomrule
    \end{tabularx}
    \medskip
    \footnotesize $\dagger$ $n=17$ for Gold and $n=11$ for Silver/Bronze.
    \label{tab:dataset_features}
\end{center}
\end{table}

%% file: Table/Gold_taxonomy.tex
\begin{table*}[!ht]
            \caption{Class definitions for the 3DSES dataset.}
            \label{tab:classes}
            \begin{tabularx}{\textwidth}{llX}
            \toprule
            \textbf{Index} & \textbf{Class name} & 
            \textbf{Definition}\\
            \midrule
            0 & Column & vertical structural element that supports weight, typically made of concrete, or metal.\\
            1 & Components & building equipment excluding furniture (\eg trash bin, hotspot wifi, electronics, \etc) \\
            2 & Covering & upper interior element of a room (\eg suspended ceiling) \\
            3 & Damper & smoke detectors \\
            4 & Door & moving building element that provides access for people to pass through\\
            5 & Exit sign & building element that indicates emergency exit, typically with green lighting \\
            6 & Fire terminal & building equipment for fire safety, that provides fluid to suppress fire or that triggers audible alarms\\
            7 & Furniture & Common furnishings such as chairs, tables, \etc \\
            8 & Heater & building element that provides heat, includes the pipes \\
            9 & Lamp & building element that provides artificial light \\
            10 & Outlet & utility element that provides access to electrical power \\
            11 & Railing & frame assembly adjacent to some boundaries or human circulations (\eg stairs) \\
            12 & Slab & structural element providing the lower support (often made in concrete) \\
            13 & Stair & structural element that allows moving between floors \\
            14 & Switch & utility element that controls the flow of electricity, typically to a lamp\\
            15 & Wall & vertical structural element, often made of stone or concrete, that divides or encloses a space\\
            16 & Window & building element that provides natural light and/or fresh air\\
            17 & Clutter & all elements that are unrelated to the building structure and equipment, \eg clothes, plants, small office supplies, persons, \etc\\
            \bottomrule
            \end{tabularx}
\end{table*}

%% file: Suppmaterial/Image_maquette.tex
\begin{figure*}[ht]

    \begin{subfigure}[b]{0.33\textwidth}
        \centering
        \includegraphics[height=3cm]{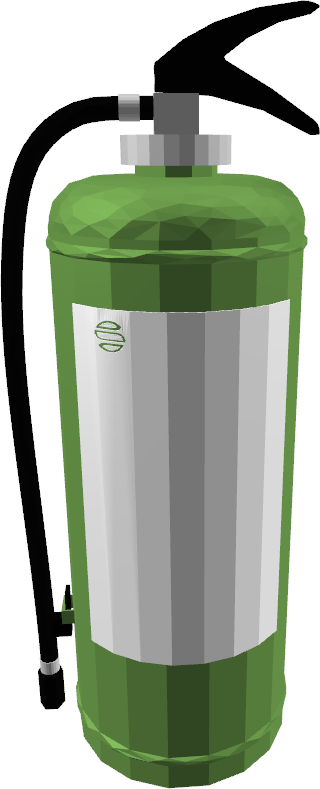}
        \caption{Extinguisher}
    \end{subfigure}%
    \hfill
    \begin{subfigure}[b]{0.33\textwidth}
        \centering
        \includegraphics[height=3cm]{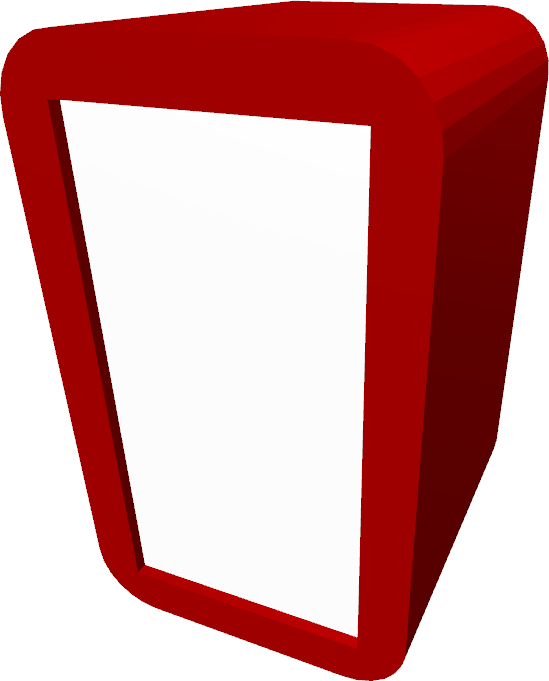}
        \caption{Manual fire alarm}
    \end{subfigure}%
    \hfill
    \begin{subfigure}[b]{0.33\textwidth}
        \centering
        \includegraphics[height=3cm]{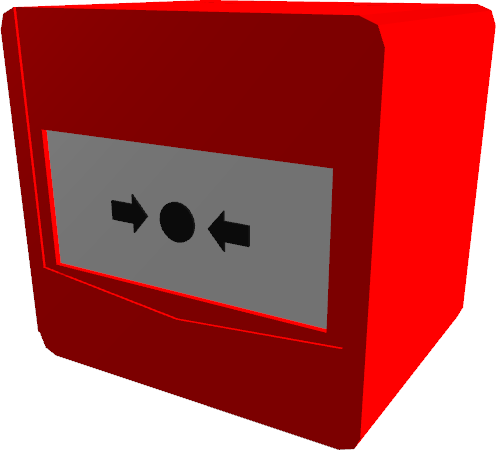}
        \caption{Manual fire alarm}
    \end{subfigure}
    \caption{Different objects from the ``Fire terminal'' class.}
    \label{fire_terminal}

    \begin{subfigure}[b]{0.33\textwidth}
        \centering
        \includegraphics[height=2cm]{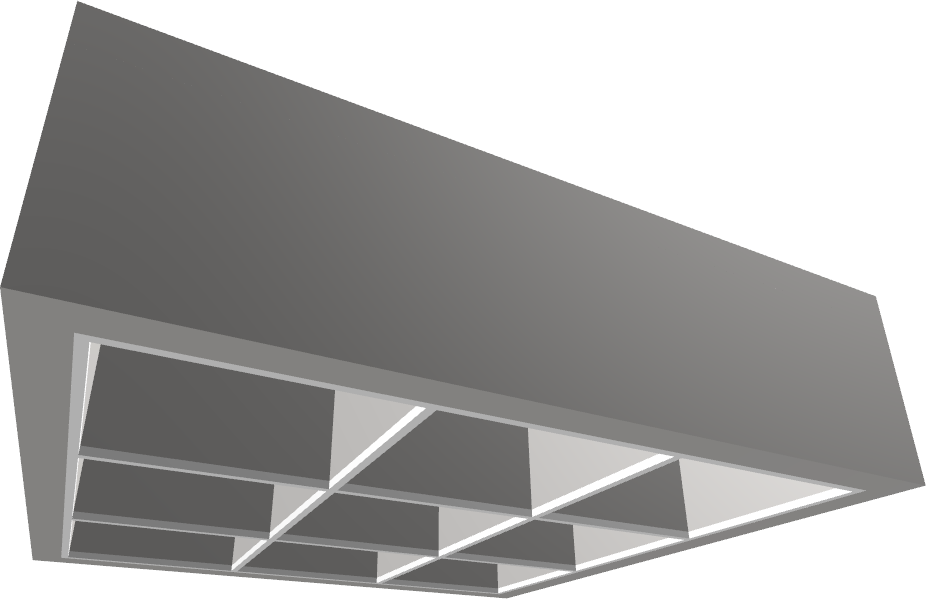}
        \caption{Lamp}
    \end{subfigure}%
    \hfill
    \begin{subfigure}[b]{0.33\textwidth}
        \centering
        \includegraphics[height=2cm]{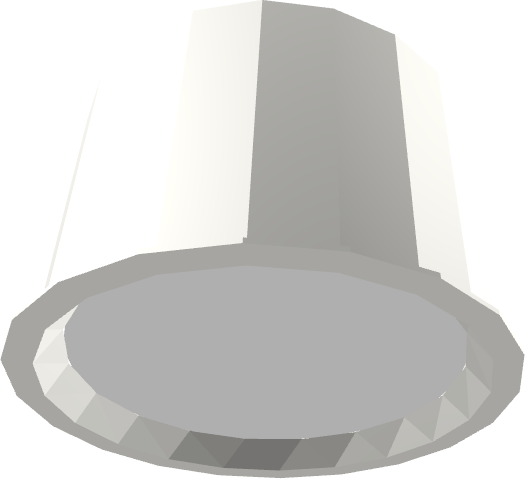}
        \caption{Lamp}
    \end{subfigure}%
    \hfill
    \begin{subfigure}[b]{0.33\textwidth}
        \centering
        \includegraphics[height=2cm]{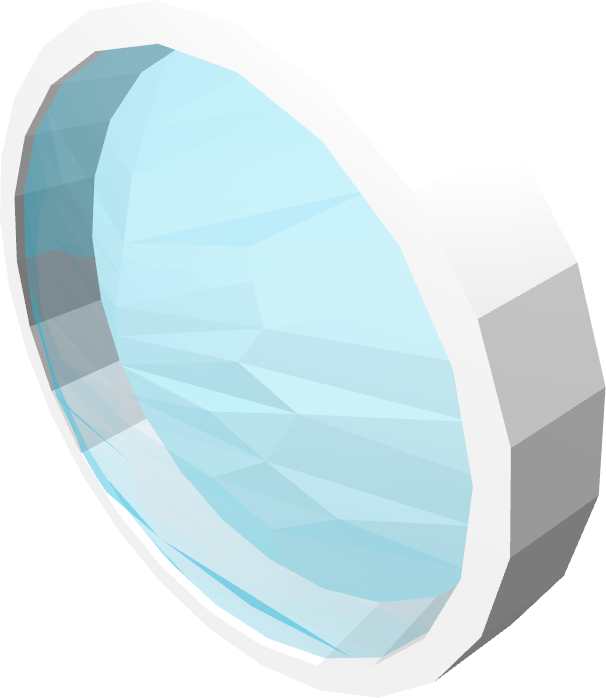}
        \caption{Lamp}
    \end{subfigure}
    \caption{Different models for the ``Lamp'' class.}
    \label{lamp}

    \begin{subfigure}[b]{0.33\textwidth}
        \centering
        \includegraphics[height=3cm]{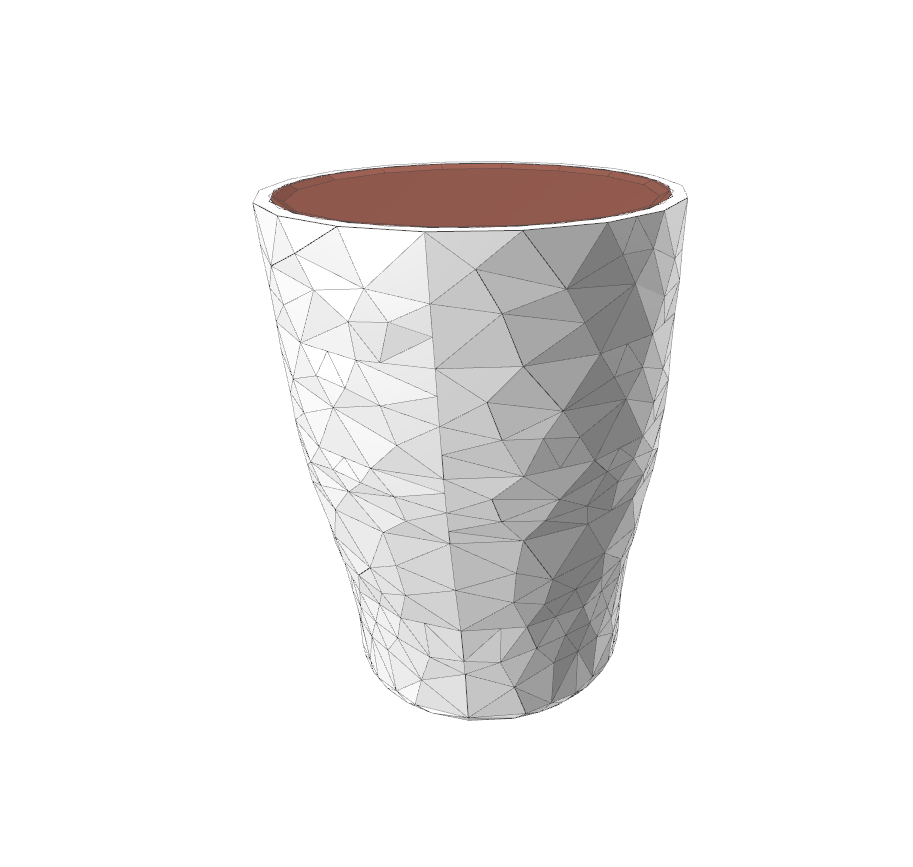}
        \caption{Trash bin}
    \end{subfigure}%
    \hfill
    \begin{subfigure}[b]{0.33\textwidth}
        \centering
        \includegraphics[height=3cm]{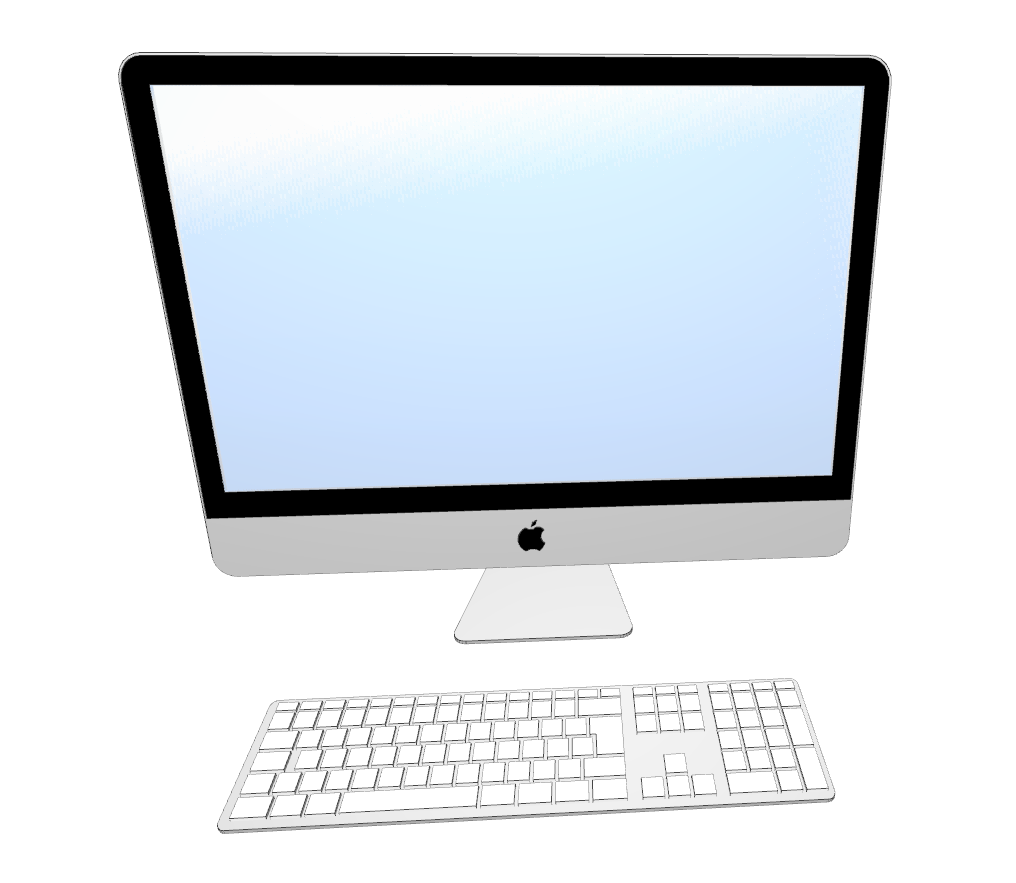}
        \caption{Screen monitor}
    \end{subfigure}%
    \hfill
    \begin{subfigure}[b]{0.33\textwidth}
        \centering
        \includegraphics[height=3cm]{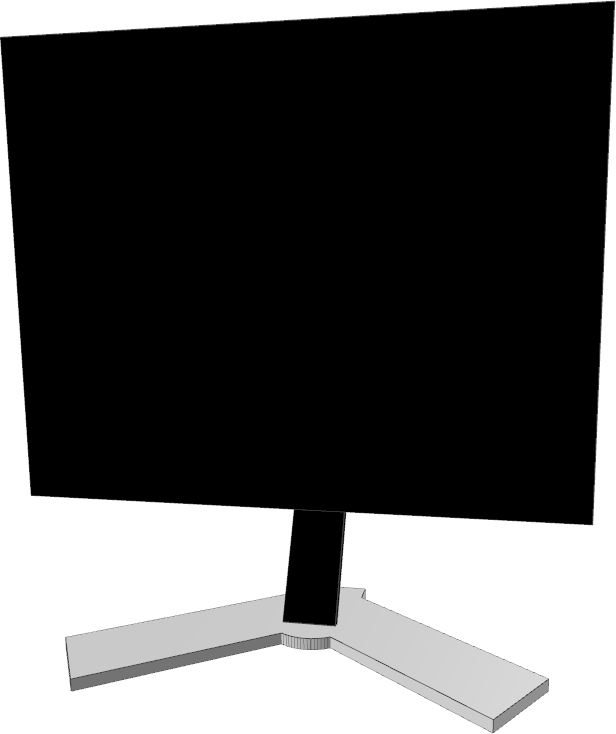}
        \caption{Screen monitor}
    \end{subfigure}
    \caption{Examples of objects from the ``Components'' class.}
    \label{components}

    \begin{subfigure}[b]{0.33\textwidth}
        \centering
        \includegraphics[height=2cm]{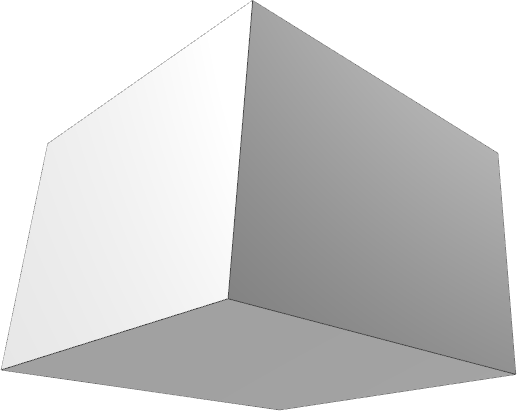}
        \caption{Hotspot (``Components'')}
    \end{subfigure}%
    \begin{subfigure}[b]{0.33\textwidth}
        \centering
        \includegraphics[height=2cm]{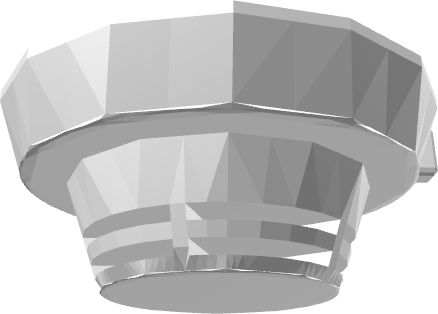}
        \caption{Smoke detector ``Damper''}
    \end{subfigure}%
    \hfill
    \begin{subfigure}[b]{0.33\textwidth}
        \centering
        \includegraphics[height=2cm]{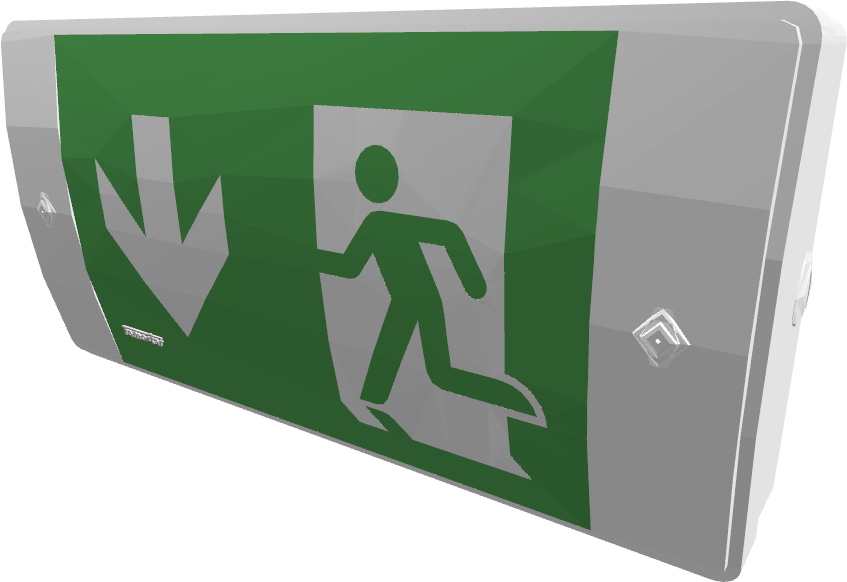}
        \caption{``Exit sign''}
    \end{subfigure}
    \caption{Example of domain specific objects from ``Components'', ``Damper'' and ``Exit sign'' classes.}
    \label{other}

    \begin{subfigure}[b]{0.33\textwidth}
        \centering
        \includegraphics[height=3cm]{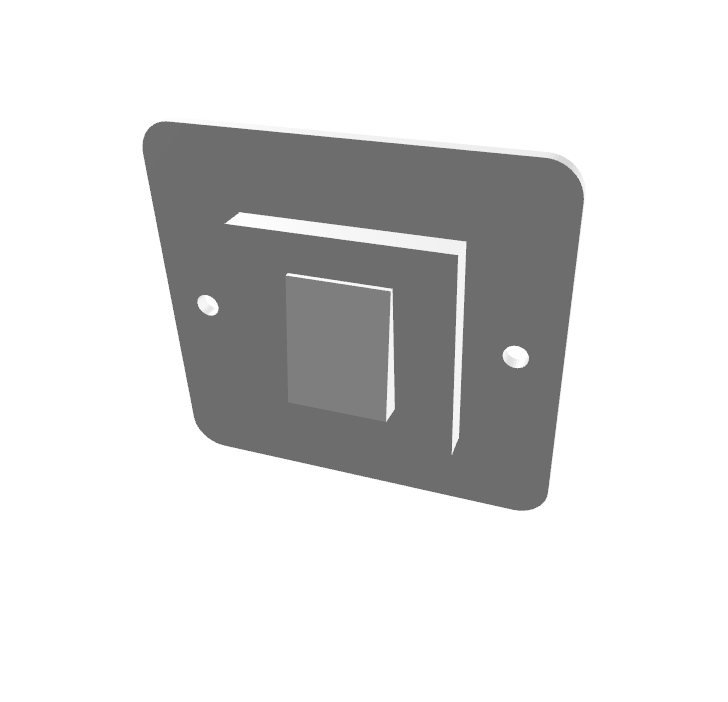}
        \caption{``Switch''}
    \end{subfigure}%
    \begin{subfigure}[b]{0.33\textwidth}
        \centering
        \includegraphics[height=3cm]{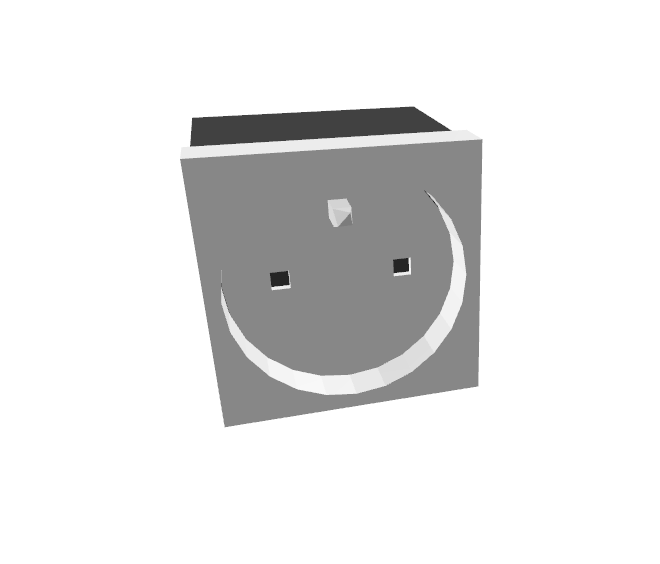}
        \caption{``Outlet''}
    \end{subfigure}%
    \hfill
    \begin{subfigure}[b]{0.33\textwidth}
        \centering
        \includegraphics[height=3.5cm]{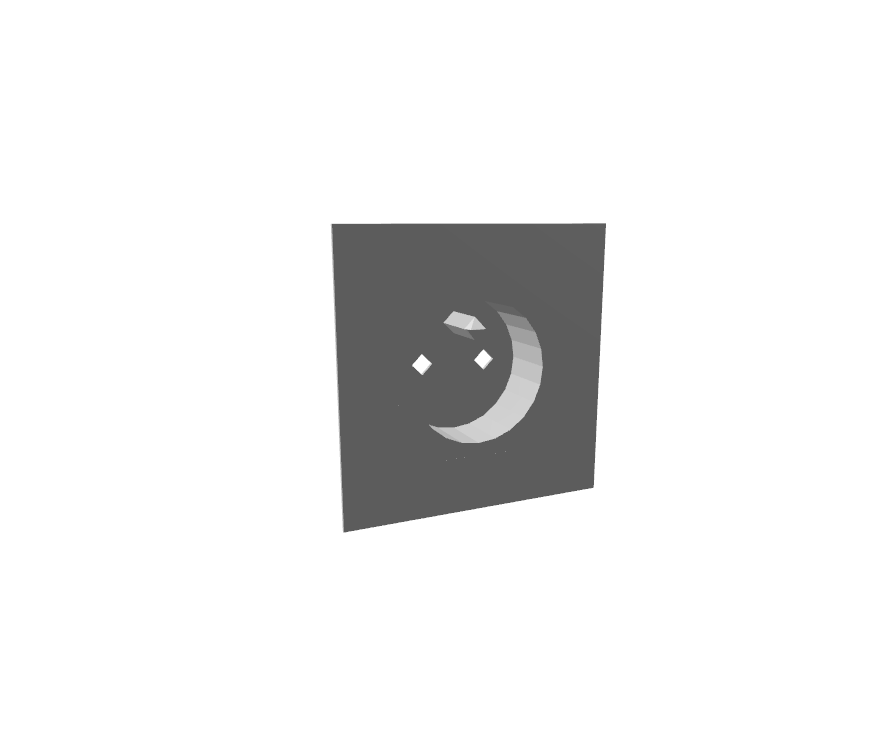}
        \caption{``Outlet''}
    \end{subfigure}
    \caption{Example of domain specific objects from ``Switch'' and ``Outlet'' classes.}
    \label{switch_and_outlet}

    \begin{subfigure}[b]{0.33\textwidth}
        \centering
        \includegraphics[height=2.5cm]{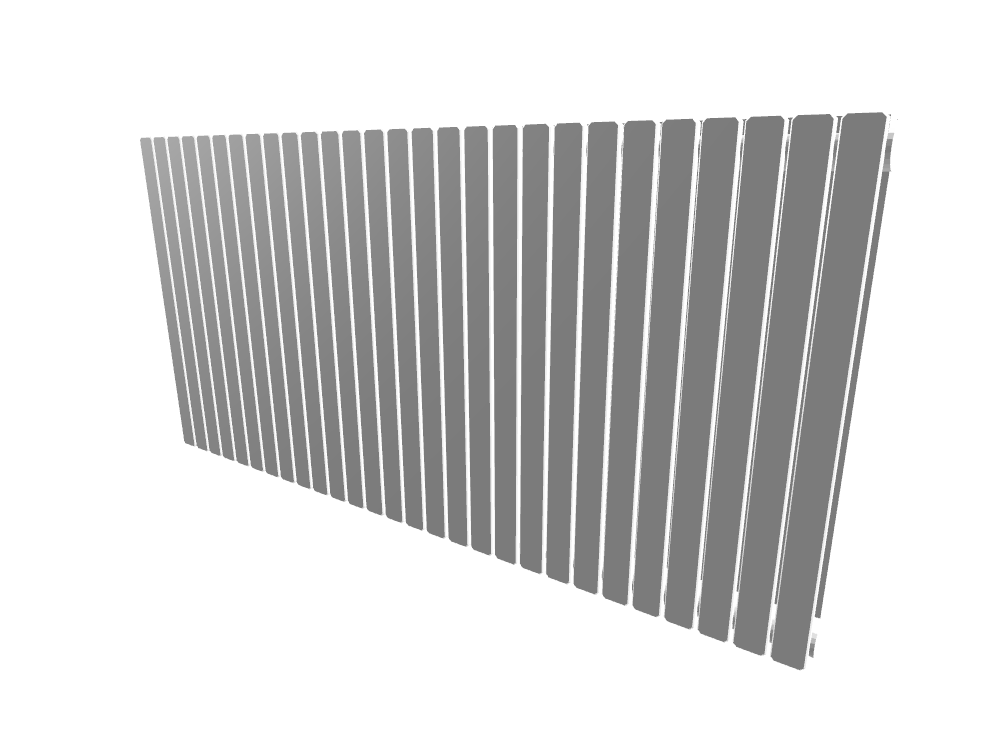}
        \caption{Radiator}
    \end{subfigure}%
    \begin{subfigure}[b]{0.33\textwidth}
        \centering
        \includegraphics[height=2.2cm]{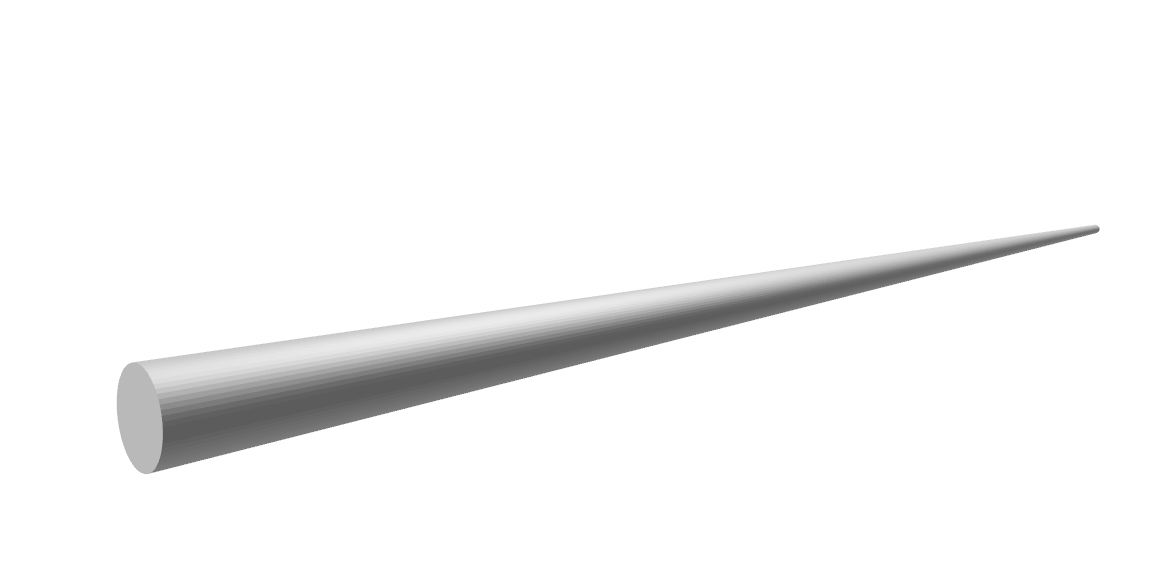}
        \caption{Pipe segment}
    \end{subfigure}%
    \hfill
    \begin{subfigure}[b]{0.33\textwidth}
        \centering
        \includegraphics[height=2.2cm]{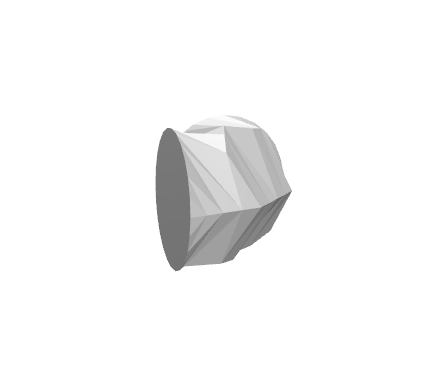}
        \caption{Pipe fitting}
    \end{subfigure}
    \caption{Example of objects from ``Heater'' class.}
    \label{heater}

\end{figure*}

%% file: Suppmaterial/Image_furniture.tex
\begin{figure*}[ht]

    \begin{subfigure}[b]{0.33\textwidth}
        \centering
        \includegraphics[height=3cm]{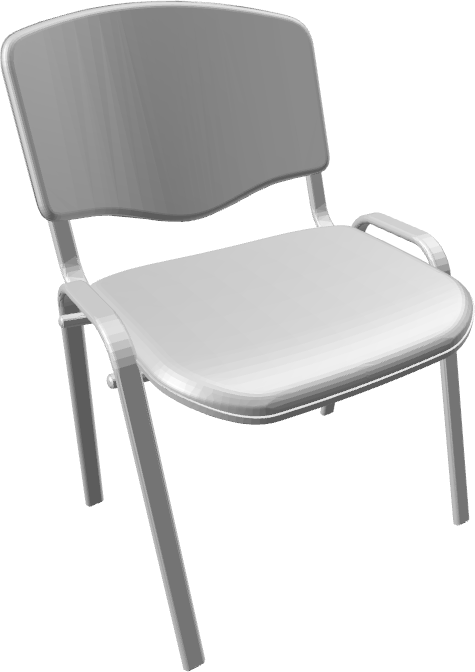}
        \caption{Chair}
    \end{subfigure}%
    \hfill
    \begin{subfigure}[b]{0.33\textwidth}
        \centering
        \includegraphics[height=3cm]{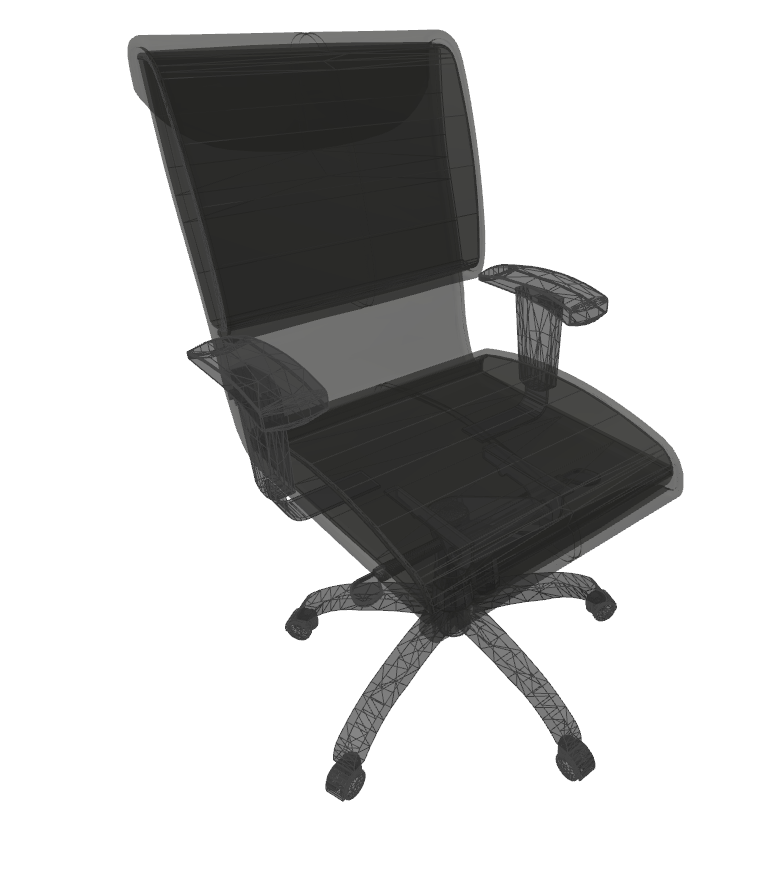}
        \caption{Chair}
    \end{subfigure}%
    \hfill
    \begin{subfigure}[b]{0.33\textwidth}
        \centering
        \includegraphics[height=3cm]{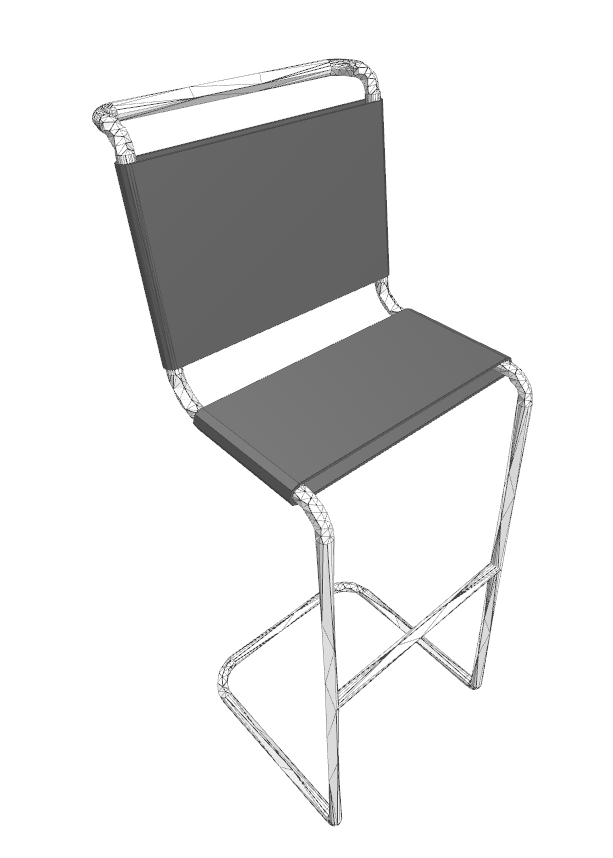}
        \caption{Chair}
    \end{subfigure}

    \begin{subfigure}[b]{0.33\textwidth}
        \centering
        \includegraphics[height=3cm]{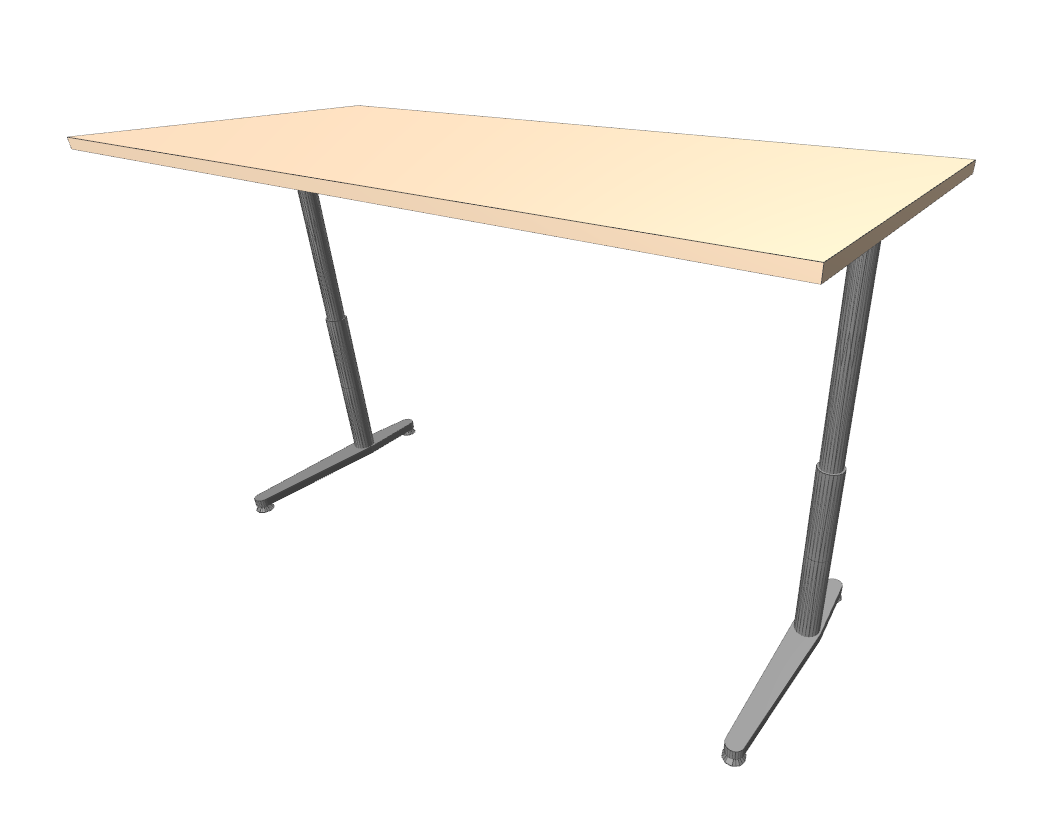}
        \caption{Desk}
    \end{subfigure}%
    \hfill
    \begin{subfigure}[b]{0.33\textwidth}
        \centering
        \includegraphics[height=3cm]{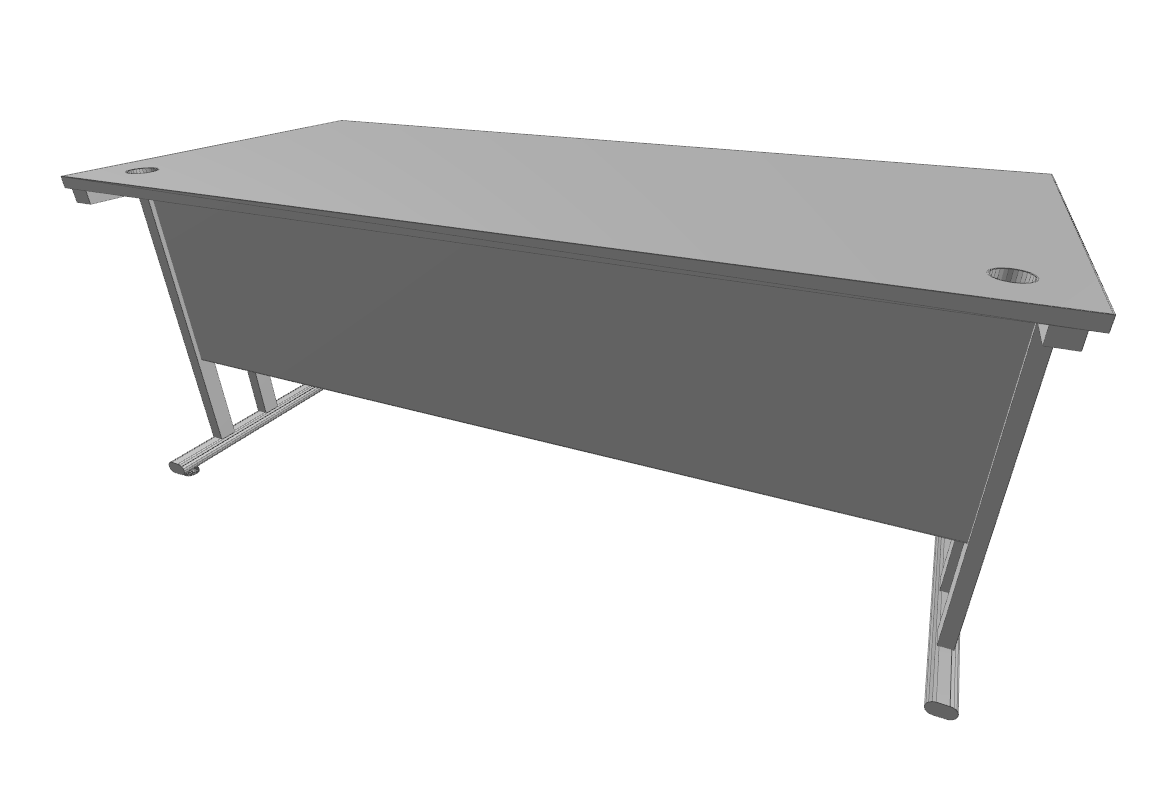}
        \caption{Desk}
    \end{subfigure}%
    \hfill
    \begin{subfigure}[b]{0.25\textwidth}
        \centering
        \includegraphics[height=3cm]{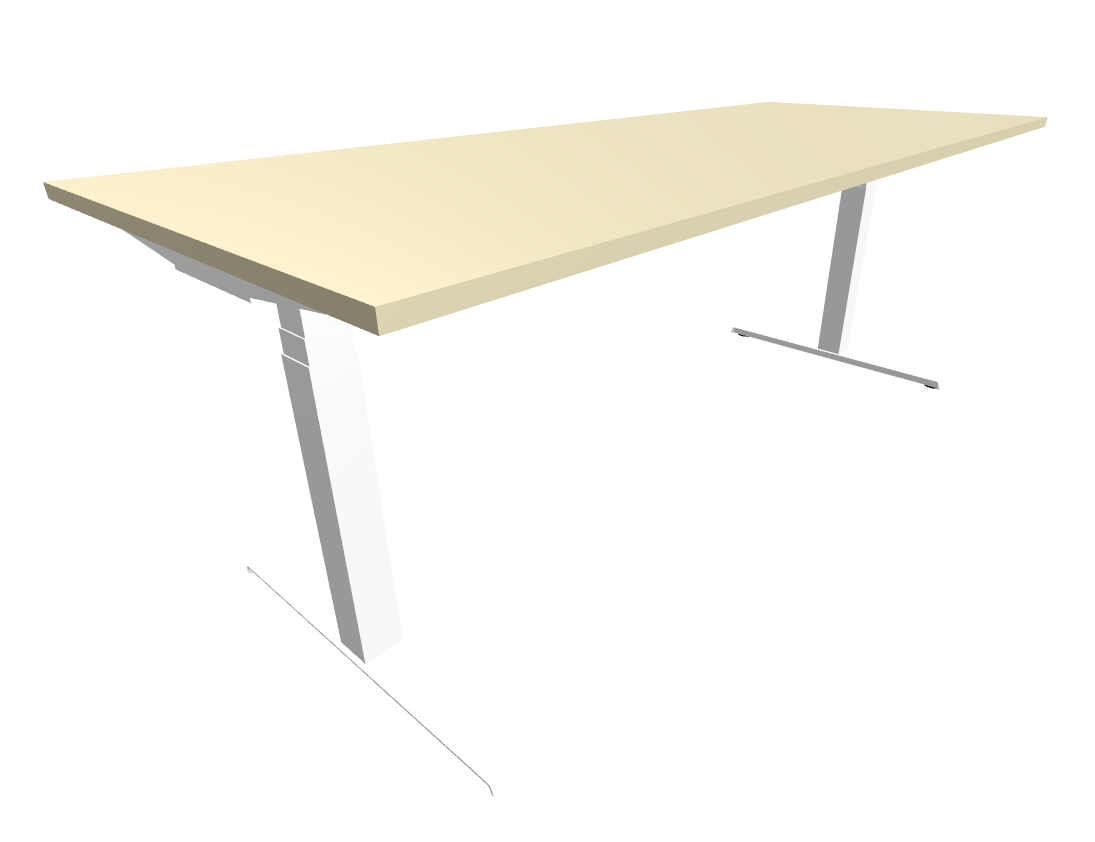}
        \caption{Desk}
    \end{subfigure}

    \begin{subfigure}[b]{0.33\textwidth}
        \centering
        \includegraphics[height=3cm]{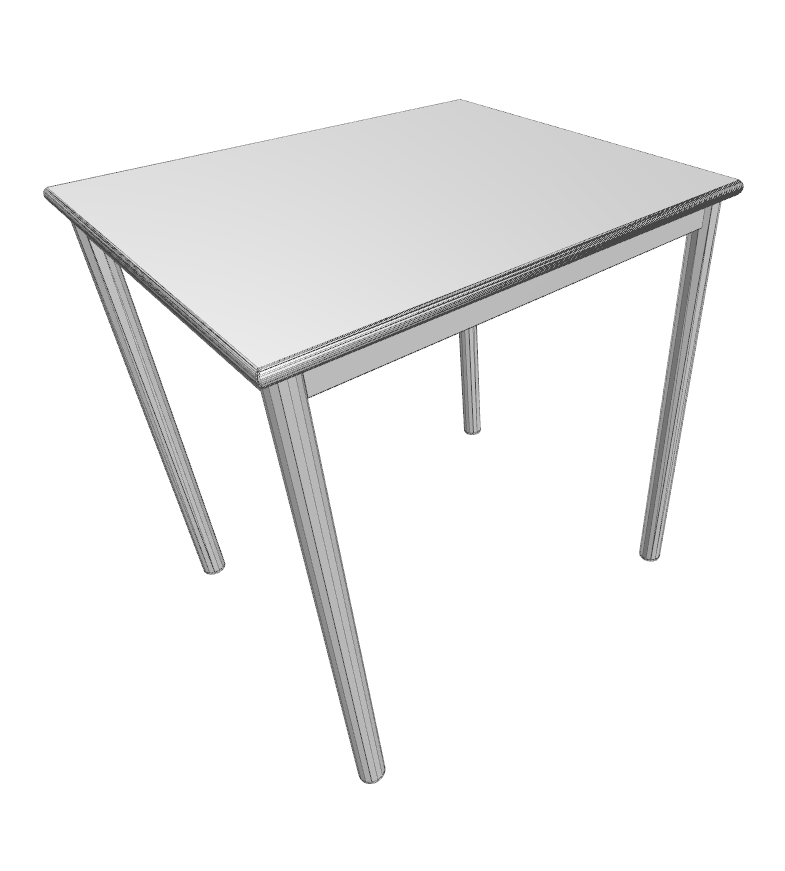}
        \caption{Table}
    \end{subfigure}%
    \hfill
    \begin{subfigure}[b]{0.33\textwidth}
        \centering
        \includegraphics[height=3cm]{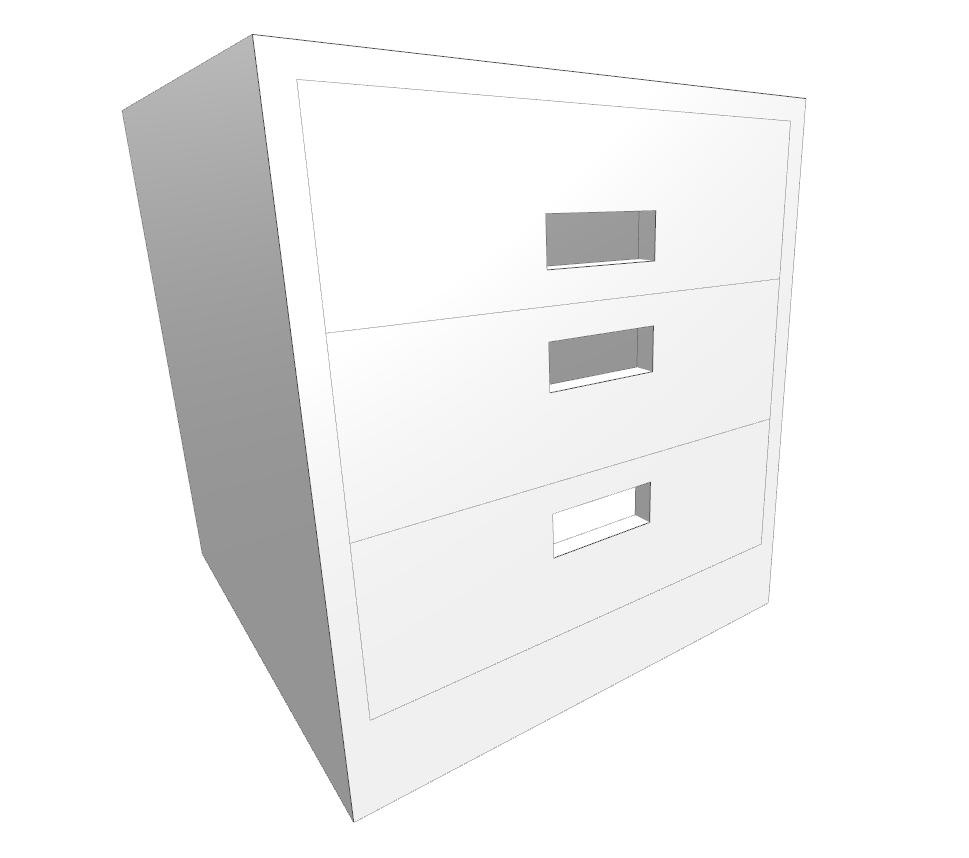}
        \caption{File cabinet}
    \end{subfigure}%
    \hfill
    \begin{subfigure}[b]{0.33\textwidth}
        \centering
        \includegraphics[height=3cm]{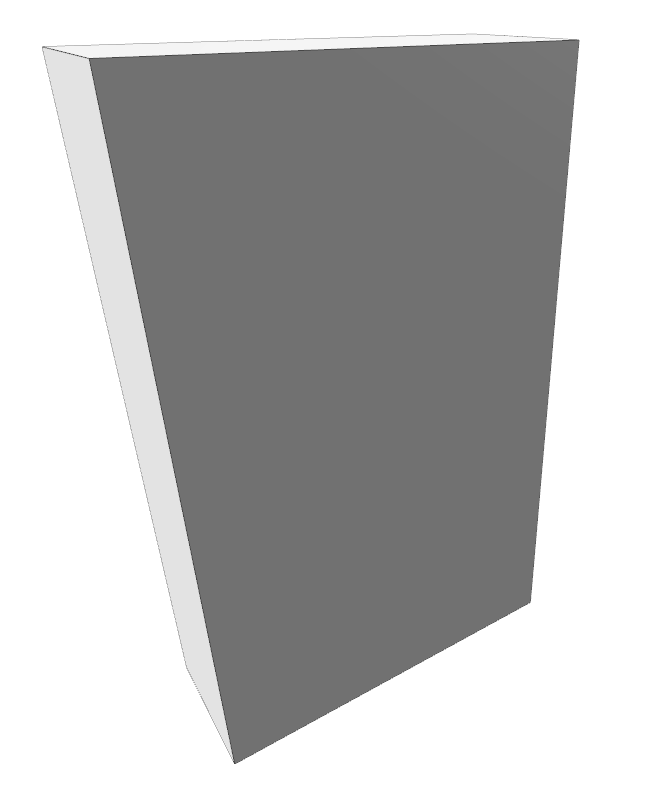}
        \caption{File cabinet}
    \end{subfigure}
    \caption{Examples of ``Furniture'' objects}
    \label{furniture}

\end{figure*}

%% file: Table/Silver_taxonomy.tex
\begin{table*}[!ht]
    \caption{Simplified class definitions for the 3DSES dataset.}
    \label{tab:classes_simplified}
    \begin{tabularx}{\textwidth}{llX}
    \toprule
    \textbf{Index} & \textbf{Class name} & 
    \textbf{Definition}\\
    \midrule
    0 & Column & vertical structural element that supports weight, typically made of concrete, or metal.\\
    1 & Covering & upper interior element of a room (\eg suspended ceiling) \\
    2 & Door & moving building element that provides access for people to pass through\\
    3 & Exit sign & building element that indicates emergency exit, typically with green lighting \\
    4 & Heater & building element that provides heat, includes the pipes \\
    5 & Lamp & building element that provides artificial light \\
    6 & Railing & frame assembly adjacent to some boundaries or human circulations (\eg stairs) \\
    7 & Slab & structural element providing the lower support (often made in concrete) \\
    8 & Stair & structural element that allows moving between floors \\
    9 & Wall & vertical structural element, often made of stone or concrete, that divides or encloses a space\\
    10 & Window & building element that provides natural light and/or fresh air\\
    11 & Clutter & all elements that are unrelated to the building structure and equipment, \eg clothes, plants, small office supplies, persons, \etc\\
        \bottomrule
    \end{tabularx}
\end{table*}

%% file: Algo/Algo.tex
\begin{algorithm}[ht]
\caption{Pseudo-labeling of the point cloud using model-to-cloud alignment.}
\label{algo2}
\SetAlgoLined
\SetKwComment{Comment}{/* }{ */}

\KwData{$\text{point cloud} \gets [p_1, p_2, \dots, p_n]$ \Comment*[r]{list of (x,y,z) points}}
\KwData{$\text{model} \gets [\text{mesh}_1, \text{mesh}_2, \dots, \text{mesh}_k]$ \Comment*[r]{list of 3D objects}}
\KwData{$\tau \geq 0$ \Comment*[r]{threshold}}

    $\text{classes} \gets$ initialize list of size $n$ with ``Clutter''\;
    \For{$p$ in point cloud}{
        $d_\text{min} = +\infty$\;
        \For{$\text{mesh}$ in $\text{model}$}{
            $d \gets$ compute signed distance between $p = (x,y,z)$ and mesh\;
            \uIf{$d \leq 0$\Comment*[r]{\footnotesize point inside the object}}{
                classes[$p$] $\gets$ class(mesh)\;
            }
            \uElseIf{$d \leq \tau$ \textbf{and} $d < d_\text{min}$ \Comment*[r]{\scriptsize p outside but near the closest object}}{
                    classes[$p$] $\gets$ class(mesh)\;
                    $d_\text{min} = d$\;
                }
            }
        }
    
    \KwResult{classes \Comment*[r]{\scriptsize classified point cloud}}

\end{algorithm}

%% file: Histograms/Computation.tex
\definecolor{goldenyellow}{rgb}{1.0, 0.84, 0.0}

\begin{figure*}
\begin{center}
\begin{tikzpicture}
  \begin{axis}[title  = Computation Time (s) - Gold \emoji{1st-place-medal} Version,
    xbar,
    height = 9.5cm,
    y axis line style = { opacity = 0 },
    ytick             = data,
    typeset ticklabels with strut, %
    xlabel=Time in seconds,
    tickwidth         = 0pt,
    enlarge y limits  = 0.05,
    enlarge x limits  = 0.02,
    symbolic y coords = {Window, Wall, Switch, Stair, Slab, Railing, Outlet, Lamp, Heater, Furniture, FireTerminal, Exit sign, Door, Damper, Covering, Components, Column},
    nodes near coords,  
  ]
 
  \addplot [fill=goldenyellow] coordinates { (7.65,Column)         (32.41,Components)
                         (7.40,Covering)       (55.39,Damper)
                         (18.12,Door)          (178.18,Exit sign)
                         (91.36,FireTerminal)  (1871.41,Furniture)
                         (132.85,Heater)        (23.52,Lamp)
                         (24.21,Outlet)       (30.74,Railing)
                         (1.36,Slab)           (7.77,Stair)
                         (14.00,Switch)          (7.31,Wall)
                         (20.27,Window)
                         };   
  \legend{Classes} %
  \end{axis}
\end{tikzpicture}

\caption{Computation time in seconds for each classes of Gold version}
\label{histogramm_time}
\end{center}
\end{figure*}
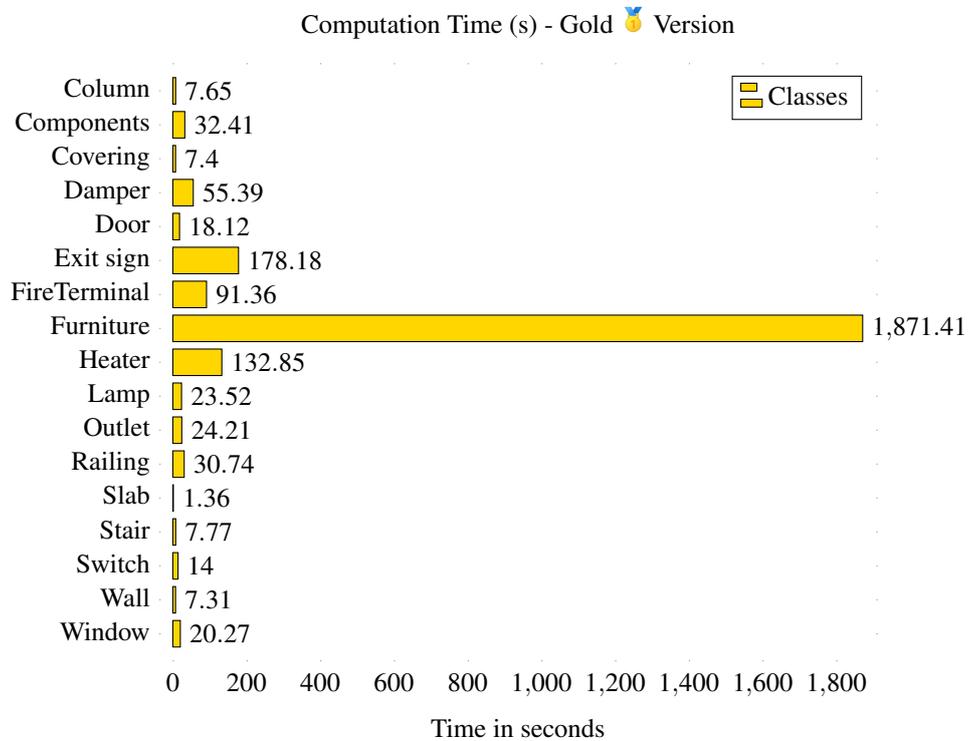

%% file: Confusionmatrix/Confusion_matrix_Gold.tex
\begin{figure*}[ht!]
    \centering
    \label{cm_gold}

    \begin{tikzpicture}[scale=0.75]
    \begin{axis}[
            colormap={gold_yellow}{rgb255=(255,255,255) rgb255=(255,184,10)},
            width=14cm,
            height=14cm,
            xlabel=Predicted,
            xlabel style={yshift=0pt},
            ylabel=Ground Truth,
            ylabel style={yshift=0pt},
            xticklabels={Column, Components, Covering, Damper, Door, Exit sign, FireTerminal, Furniture, Heater, Lamp, Outlet, Railing, Slab, Stair, Switch, Wall, Window, Clutter},
            xtick={0,...,17},
            xtick style={draw=none},
            yticklabels={Column, Components, Covering, Damper, Door, Exit sign, FireTerminal, Furniture, Heater, Lamp, Outlet, Railing, Slab, Stair, Switch, Wall, Window, Clutter},
            ytick={0,...,17},
            ytick style={draw=none},
            enlargelimits=false,
            xticklabel style={
              rotate=90
            },
            colorbar,
            nodes near coords={\pgfmathprintnumber\Cvalue},
            visualization depends on={\thisrow{C} \as \Cvalue},
            nodes near coords style={
                yshift=-7pt
            },
        ]
        \addplot[
            matrix plot,
            mesh/cols=18,
            point meta={\thisrow{C}},
            draw=gray
        ] table {
            x y C
            0 0 97.3
            1 0 0.0
            2 0 0.2
            3 0 0.0
            4 0 0.0
            5 0 0.0
            6 0 0.0
            7 0 0.0
            8 0 0.0
            9 0 0.0
            10 0 0.0
            11 0 0.0
            12 0 0.0
            13 0 0.0
            14 0 0.0
            15 0 2.5
            16 0 0.0
            17 0 0.0

            0 1 0.0
            1 1 84.9
            2 1 0.0
            3 1 0.0
            4 1 0.0
            5 1 0.0
            6 1 0.0
            7 1 2.0
            8 1 0.0
            9 1 0.0
            10 1 0.0
            11 1 0.0
            12 1 1.5
            13 1 0.0
            14 1 0.0
            15 1 0.3
            16 1 0.0
            17 1 11.4

            0 2 0.0
            1 2 0.0
            2 2 96.7
            3 2 0.0
            4 2 0.0
            5 2 0.0
            6 2 0.0
            7 2 0.0
            8 2 0.0
            9 2 0.1
            10 2 0.0
            11 2 0.0
            12 2 0.0
            13 2 1.0
            14 2 0.0
            15 2 1.3
            16 2 0.0
            17 2 0.9

            0 3 0.0
            1 3 0.0
            2 3 12.5
            3 3 87.5
            4 3 0.0
            5 3 0.0
            6 3 0.0
            7 3 0.0
            8 3 0.0
            9 3 0.0
            10 3 0.0
            11 3 0.0
            12 3 0.0
            13 3 0.0
            14 3 0.0
            15 3 0.0
            16 3 0.0
            17 3 0.0

            0 4 0.0
            1 4 0.0
            2 4 0.0
            3 4 0.0
            4 4 95.5
            5 4 0.0
            6 4 0.0
            7 4 0.0
            8 4 0.0
            9 4 0.0
            10 4 0.0
            11 4 0.0
            12 4 0.3
            13 4 0.0
            14 4 0.0
            15 4 3.3
            16 4 0.6
            17 4 0.3

            0 5 0.0
            1 5 0.0
            2 5 0.0
            3 5 0.0
            4 5 0.0
            5 5 91.3
            6 5 0.0
            7 5 0.0
            8 5 0.0
            9 5 0.0
            10 5 0.0
            11 5 0.0
            12 5 0.0
            13 5 0.0
            14 5 0.0
            15 5 8.7
            16 5 0.0
            17 5 0.0

            0 6 0.0
            1 6 0.0
            2 6 0.0
            3 6 0.0
            4 6 0.0
            5 6 0.0
            6 6 93.6
            7 6 0.0
            8 6 0.0
            9 6 0.1
            10 6 0.0
            11 6 0.0
            12 6 0.0
            13 6 0.0
            14 6 0.0
            15 6 1.7
            16 6 0.0
            17 6 4.7

            0 7 0.0
            1 7 0.2
            2 7 0.0
            3 7 0.0
            4 7 0.0
            5 7 0.0
            6 7 0.0
            7 7 85.6
            8 7 0.0
            9 7 0.0
            10 7 0.0
            11 7 0.0
            12 7 0.9
            13 7 0.0
            14 7 0.0
            15 7 1.2
            16 7 0.0
            17 7 12.2

            0 8 0.0
            1 8 0.0
            2 8 0.0
            3 8 0.0
            4 8 0.0
            5 8 0.0
            6 8 0.3
            7 8 0.0
            8 8 96.9
            9 8 0.0
            10 8 0.0
            11 8 0.0
            12 8 0.0
            13 8 0.0
            14 8 0.0
            15 8 2.6
            16 8 0.0
            17 8 0.2

            0 9 0.0
            1 9 0.0
            2 9 16.9
            3 9 0.1
            4 9 0.0
            5 9 0.0
            6 9 0.0
            7 9 0.0
            8 9 0.0
            9 9 69.1
            10 9 0.0
            11 9 0.0
            12 9 0.0
            13 9 0.0
            14 9 0.0
            15 9 2.1
            16 9 0.0
            17 9 11.8

            0 10 0.0
            1 10 0.0
            2 10 0.0
            3 10 0.0
            4 10 0.0
            5 10 0.0
            6 10 0.0
            7 10 0.0
            8 10 0.0
            9 10 0.0
            10 10 41.6
            11 10 0.0
            12 10 0.0
            13 10 0.0
            14 10 0.0
            15 10 58.3
            16 10 0.0
            17 10 0.0

            0 11 0.0
            1 11 0.0
            2 11 0.0
            3 11 0.0
            4 11 0.0
            5 11 0.0
            6 11 0.0
            7 11 0.0
            8 11 0.0
            9 11 0.0
            10 11 0.0
            11 11 60.6
            12 11 0.3
            13 11 2.8
            14 11 0.0
            15 11 0.9
            16 11 0.0
            17 11 35.3

            0 12 0.0
            1 12 0.0
            2 12 0.0
            3 12 0.0
            4 12 0.2
            5 12 0.0
            6 12 0.0
            7 12 0.1
            8 12 0.0
            9 12 0.0
            10 12 0.0
            11 12 0.0
            12 12 97.1
            13 12 1.8
            14 12 0.0
            15 12 0.3
            16 12 0.3
            17 12 0.1

            0 13 0.0
            1 13 0.0
            2 13 0.0
            3 13 0.0
            4 13 0.0
            5 13 0.0
            6 13 0.0
            7 13 0.0
            8 13 0.0
            9 13 0.0
            10 13 0.0
            11 13 0.1
            12 13 0.4
            13 13 97.4
            14 13 0.0
            15 13 1.2
            16 13 0.0
            17 13 0.9

            0 14 0.0
            1 14 0.0
            2 14 0.0
            3 14 0.0
            4 14 0.0
            5 14 0.0
            6 14 0.0
            7 14 0.0
            8 14 0.0
            9 14 0.0
            10 14 0.0
            11 14 0.0
            12 14 0.0
            13 14 0.0
            14 14 49.8
            15 14 50.2
            16 14 0.0
            17 14 0.0

            0 15 0.2
            1 15 0.0
            2 15 0.3
            3 15 0.0
            4 15 0.4
            5 15 0.0
            6 15 0.0
            7 15 0.1
            8 15 0.1
            9 15 0.0
            10 15 0.0
            11 15 0.0
            12 15 0.3
            13 15 0.0
            14 15 0.0
            15 15 97.8
            16 15 0.3
            17 15 0.2

            0 16 0.0
            1 16 0.0
            2 16 0.3
            3 16 0.0
            4 16 5.5
            5 16 0.0
            6 16 0.0
            7 16 0.0
            8 16 0.0
            9 16 0.0
            10 16 0.0
            11 16 0.0
            12 16 0.7
            13 16 0.0
            14 16 0.0
            15 16 22.6
            16 16 70.4
            17 16 0.5

            0 17 0.1
            1 17 0.8
            2 17 0.0
            3 17 0.0
            4 17 0.0
            5 17 0.0
            6 17 0.0
            7 17 24.1
            8 17 0.2
            9 17 0.0
            10 17 0.0
            11 17 0.0
            12 17 7.0
            13 17 0.0
            14 17 0.0
            15 17 3.5
            16 17 0.5
            17 17 63.7

        };
    \end{axis}
\end{tikzpicture}

\caption{Confusion matrix for Gold Version \emoji{1st-place-medal}}
\label{tab:cm_gold}
\end{figure*}
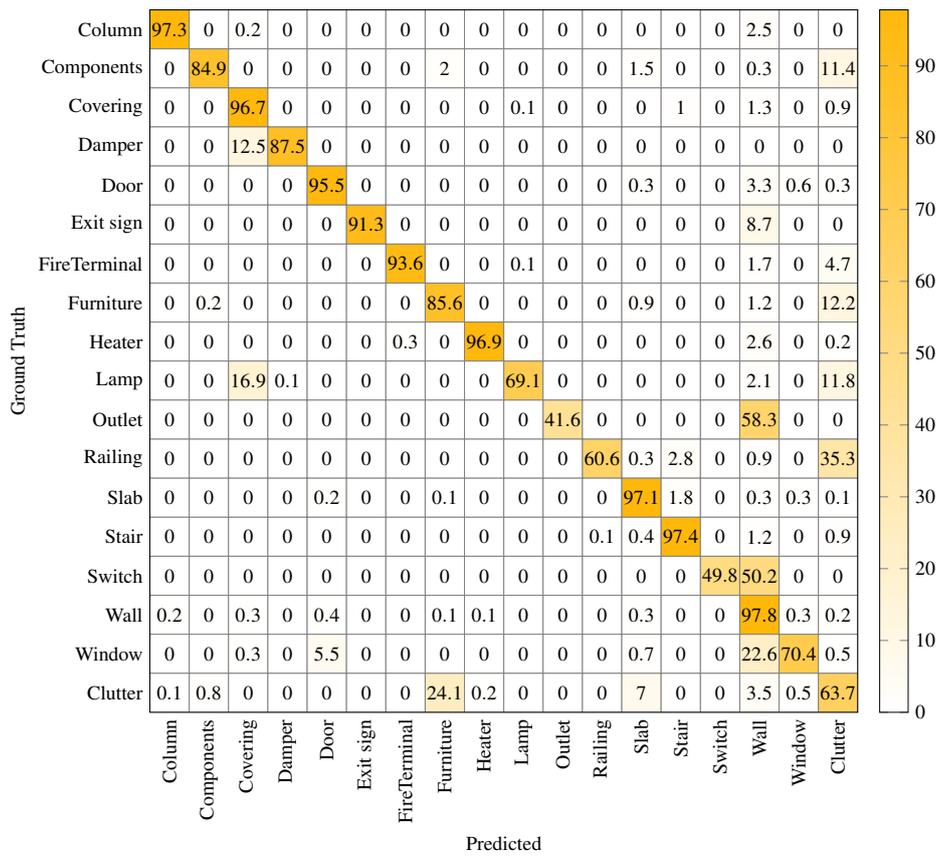

%% file: Confusionmatrix/Confusion_matrix_Silver.tex
\begin{figure*}[!ht]
    \centering
    \label{cm_silver}

    \begin{tikzpicture}[scale=0.75]
    \begin{axis}[
            colormap={gold_yellow}{rgb255=(255,255,255) rgb255=(191,191,191)},
            width=12cm,
            height=12cm,
            xlabel=Predicted,
            xlabel style={yshift=0pt},
            ylabel=Ground Truth,
            ylabel style={yshift=0pt},
            xticklabels={Column, Covering, Door, Exit sign, Heater, Lamp, Railing, Slab, Stair, Wall, Window, Clutter},
            xtick={0,...,11},
            xtick style={draw=none},
            yticklabels={Column, Covering, Door, Exit sign, Heater, Lamp, Railing, Slab, Stair, Wall, Window, Clutter},
            ytick={0,...,11},
            ytick style={draw=none},
            enlargelimits=false,
            xticklabel style={
              rotate=90
            },
            colorbar,
            nodes near coords={\pgfmathprintnumber\Cvalue},
            visualization depends on={\thisrow{C} \as \Cvalue},
            nodes near coords style={
                yshift=-7pt
            },
        ]
        \addplot[
            matrix plot,
            mesh/cols=12,
            point meta={\thisrow{C}},
            draw=gray
        ] table {
            x y C
            0 0 81.5
            1 0 5.9
            2 0 0.0
            3 0 0.0
            4 0 0.0
            5 0 0.0
            6 0 0.0
            7 0 0.0
            8 0 0.0
            9 0 3.9
            10 0 0.0
            11 0 8.8

            0 1 0.0
            1 1 98.8
            2 1 0.0
            3 1 0.0
            4 1 0.0
            5 1 0.1
            6 1 0.0
            7 1 0.0
            8 1 0.3
            9 1 0.6
            10 1 0.0
            11 1 0.3

            0 2 0.0
            1 2 0.0
            2 2 95.8
            3 2 0.0
            4 2 0.0
            5 2 0.0
            6 2 0.0
            7 2 0.2
            8 2 0.0
            9 2 2.3
            10 2 1.3
            11 2 0.4

            0 3 0.0
            1 3 0.0
            2 3 0.0
            3 3 90.9
            4 3 0.0
            5 3 0.0
            6 3 0.0
            7 3 0.0
            8 3 0.0
            9 3 9.0
            10 3 0.0
            11 3 0.1

            0 4 0.0
            1 4 0.0
            2 4 0.0
            3 4 0.0
            4 4 96.0
            5 4 0.0
            6 4 0.0
            7 4 0.0
            8 4 0.0
            9 4 2.3
            10 4 0.2
            11 4 1.5

            0 5 0.0
            1 5 12.9
            2 5 0.0
            3 5 0.0
            4 5 0.0
            5 5 76.2
            6 5 0.0
            7 5 0.0
            8 5 0.0
            9 5 0.6
            10 5 0.0
            11 5 10.2

            0 6 0.0
            1 6 0.0
            2 6 0.0
            3 6 0.0
            4 6 0.0
            5 6 0.0
            6 6 60.6
            7 6 0.3
            8 6 2.8
            9 6 0.9
            10 6 0.0
            11 6 35.3

            0 7 0.0
            1 7 0.0
            2 7 0.2
            3 7 0.0
            4 7 0.0
            5 7 0.0
            6 7 0.0
            7 7 98.8
            8 7 0.6
            9 7 0.1
            10 7 0.1
            11 7 0.2

            0 8 0.0
            1 8 0.0
            2 8 0.0
            3 8 0.0
            4 8 0.0
            5 8 0.0
            6 8 0.1
            7 8 0.4
            8 8 97.4
            9 8 1.2
            10 8 0.0
            11 8 0.9

            0 9 0.4
            1 9 0.6
            2 9 0.4
            3 9 0.0
            4 9 0.1
            5 9 0.0
            6 9 0.0
            7 9 0.3
            8 9 0.0
            9 9 97.1
            10 9 0.3
            11 9 0.7

            0 10 0.0
            1 10 0.1
            2 10 6.1
            3 10 0.0
            4 10 0.0
            5 10 0.0
            6 10 0.0
            7 10 0.5
            8 10 0.0
            9 10 20.9
            10 10 69.0
            11 10 3.3

            0 11 0.0
            1 11 0.1
            2 11 0.1
            3 11 0.0
            4 11 0.2
            5 11 0.0
            6 11 0.0
            7 11 2.3
            8 11 0.0
            9 11 3.9
            10 11 1.2
            11 11 92.3

        };
    \end{axis}
\end{tikzpicture}

\caption{Confusion matrix for Silver Version \emoji{2nd-place-medal} }
\label{tab:cm_silver}
\end{figure*}
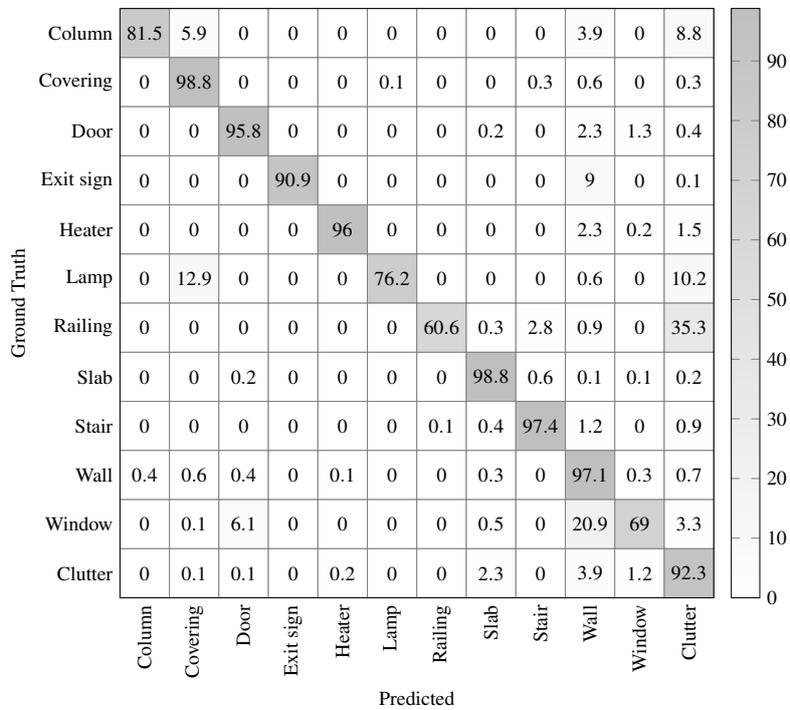

%% file: sec/X_suppl_datasheet.tex
\clearpage
\onecolumn
\section{Datasheet for 3DSES}
\label{sec:datasheet}

To help users understand the motivations and the technical characteristics of 3DSES, we provide bellow a detailed datasheet following the
\href{https://arxiv.org/abs/1803.09010}{Datasheets for Datasets} template.

\definecolor{darkblue}{RGB}{46,25, 110}

\newcommand{\dssectionheader}[1]{%
   \noindent\framebox[\columnwidth]{%
      {\fontfamily{phv}\selectfont \textbf{\textcolor{darkblue}{#1}}}
   }
}

\newcommand{\dsquestion}[1]{%
    {\noindent \fontfamily{phv}\selectfont \textcolor{darkblue}{\textbf{#1}}}
}

\newcommand{\dsquestionex}[2]{%
    {\noindent \fontfamily{phv}\selectfont \textcolor{darkblue}{\textbf{#1} #2}}
}

\newcommand{\dsanswer}[1]{%
   {\noindent #1 \medskip}
}

\dssectionheader{Motivation}

\dsquestionex{For what purpose was the dataset created?}{Was there a specific task in mind? Was there a specific gap that needed to be filled? Please provide a description.}

\dsanswer{
3DSES was created to evaluate semantic segmentation of dense indoor Terrestrial Laser Scanning (TLS) point clouds. It focuses on both structural classes (walls, floors, doors, \etc.) and building systems (\eg electrical and safety systems).
}

\dsquestion{Who created this dataset (e.g., which team, research group) and on behalf of which entity (e.g., company, institution, organization)?}

\dsanswer{
The dataset was created by the \censor{GeF (Geomatics and Land Law)} laboratory from \censor{ESGT (\emph{École supérieure des ingénieurs géomètres et topographes})}, the \censor{French engineering school of survey and topography}, located in \censor{Le Mans, France}. It was a joint work with the \censor{survey company Quarta, located in Rennes (France)} and the \censor{CEDRIC (Center for Studies and Research in Computer Science and Communication)} laboratory from \censor{Cnam (\emph{Conservatoire national des arts et métiers)}, in Paris (France)}.
}

\dsquestionex{Who funded the creation of the dataset?}{If there is an associated grant, please provide the name of the grantor and the grant name and number.}

\dsanswer{
The dataset was funded by \censor{Quarta} through a \censor{CIFRE contract} funding the \censor{Ph.D. thesis of Maxime Mérizette, the main author of the dataset}.
}

\dsquestion{Any other comments?}

\bigskip
\dssectionheader{Composition}

\dsquestionex{What do the instances that comprise the dataset represent (e.g., documents, photos, people, countries)?}{ Are there multiple types of instances (e.g., movies, users, and ratings; people and interactions between them; nodes and edges)? Please provide a description.}

\dsanswer{
The dataset is comprised of TLS scans that represent a 3D reconstruction of a part of the first floor of the main \censor{ESGT} building.
}

\dsquestion{How many instances are there in total (of each type, if appropriate)?}

\dsanswer{There are:
\begin{itemize}
    \item 10 scans in the Gold variant,
    \item 20 additional scans in the Silver variant,
    \item 12 additional scans in the Bronze variant.
\end{itemize}
Each scan contains approximately 6 million points (for Gold \& Silver version) and approximately 10 millions points (for Bronze version).

In addition, 3DSES contains a 3D CAD model of the building, tagged with objects using the standard IFC format.
}

\dsquestionex{Does the dataset contain all possible instances or is it a sample (not necessarily random) of instances from a larger set?}{ If the dataset is a sample, then what is the larger set? Is the sample representative of the larger set (e.g., geographic coverage)? If so, please describe how this representativeness was validated/verified. If it is not representative of the larger set, please describe why not (e.g., to cover a more diverse range of instances, because instances were withheld or unavailable).}

\dsanswer{
The dataset contains all scans for the first floor of the \censor{ESGT}. While it does not cover the entire building, the first floor has been entirely scanned.
3DSES is not representative of all possible indoor TLS scans of buildings, as it covers only one specific building (\censor{an engineering school}), and only its first floor.
}

\dsquestionex{What data does each instance consist of? “Raw” data (e.g., unprocessed text or images) or features?}{In either case, please provide a description.}

\dsanswer{
The TLS scans are released as colorized point clouds that describe $(x,y,z)$ coordinates in space, an RGB value and a Lidar intensity. The 3D CAD model, was composed of 3D CAD objects (with some information such as  material). For our alignment methods, all objects are merged correlated to their semantic meaning and saved to the \texttt{obj} format. Each \texttt{obj} contains a list of vertices $(x,y,z)$, a list of vertex normals $(x,y,z)$ and list of polygonal face elements (\eg a link to vertices number).
}

\dsquestionex{Is there a label or target associated with each instance?}{If so, please provide a description.}

\dsanswer{
These point clouds have been labeled in several classes (18 for the Gold variant, 12 for the Silver and Bronze variants of the dataset). One class label is given for every point in the Gold and Silver scans.

In addition, point clouds have also been pseudo-labeled using an automated algorithm. This pseudo-label uses the same 12 classes as the Silver/Bronze variants, and is available for every point, including the otherwise unlabeled points of the Bronze variant.
}

\dsquestionex{Is any information missing from individual instances?}{If so, please provide a description, explaining why this information is missing (e.g., because it was unavailable). This does not include intentionally removed information, but might include, e.g., redacted text.}

\dsanswer{
The 3D reconstruction provided by point clouds can sparse for some areas, resulting in potential ``missing data'', however this is to be expected when using Lidar scanning. Semantic labels are missing for the scans exclusive to the Bronze variants.
}

\dsquestionex{Are relationships between individual instances made explicit (e.g., users’ movie ratings, social network links)?}{If so, please describe how these relationships are made explicit.}

\dsanswer{
All scans are georeferenced, making it possible to coregister them and bundle them into a single point cloud to retrieve the full geometry of the building.
}

\dsquestionex{Are there recommended data splits (e.g., training, development/validation, testing)?}{If so, please provide a description of these splits, explaining the rationale behind them.}

\dsanswer{Yes, a predefined training and testing split is set for benchmarking on 3DSES. Three specific scans are kept private and used as a test set for evaluating methods. These scans cover a smaller area of \censor{ESGT} that contains all of the objects of the interest of the 3DSES dataset.
}

\dsquestionex{Are there any errors, sources of noise, or redundancies in the dataset?}{If so, please provide a description.}

\dsanswer{
There are two sources of errors in the dataset:
\begin{itemize}
    \item The pseudo-labels can be incorrect as they have been extracted using an automated approach. We evaluated their accuracy to be over 94\%, although this depends on the semantic class.
    \item The TLS scans used for the Bronze variant of 3DSES are raw Lidar acquisitions and can contain outliers and artefacts.    
\end{itemize}

The real labels have been manually checked by an expert in addition to the original annotation and do not contain any error to the best of our knowledge.
}

\dsquestionex{Is the dataset self-contained, or does it link to or otherwise rely on external resources (e.g., websites, tweets, other datasets)?}{If it links to or relies on external resources, a) are there guarantees that they will exist, and remain constant, over time; b) are there official archival versions of the complete dataset (i.e., including the external resources as they existed at the time the dataset was created); c) are there any restrictions (e.g., licenses, fees) associated with any of the external resources that might apply to a future user? Please provide descriptions of all external resources and any restrictions associated with them, as well as links or other access points, as appropriate.}

\dsanswer{
The 3DSES dataset is entirely self contained. The test set labels are kept hidden for the time being.
}

\dsquestionex{Does the dataset contain data that might be considered confidential (e.g., data that is protected by legal privilege or by doctor-patient confidentiality, data that includes the content of individuals non-public communications)?}{If so, please provide a description.}

\dsanswer{
There is no confidential or privileged data in the 3DSES. We have obtained the agreement of the \censor{ESGT director} to release the data.
}

\dsquestionex{Does the dataset contain data that, if viewed directly, might be offensive, insulting, threatening, or might otherwise cause anxiety?}{If so, please describe why.}

\dsanswer{
No.
}

\dsquestionex{Does the dataset relate to people?}{If not, you may skip the remaining questions in this section.}

\dsanswer{
No, all individuals present during the TLS scans were asked to move outside the acquisition range during data collection. As a result, no humans are visible in the scans.
}

\dsquestionex{Does the dataset identify any subpopulations (e.g., by age, gender)?}{If so, please describe how these subpopulations are identified and provide a description of their respective distributions within the dataset.}

\dsanswer{
No.
}

\dsquestionex{Is it possible to identify individuals (i.e., one or more natural persons), either directly or indirectly (i.e., in combination with other data) from the dataset?}{If so, please describe how.}

\dsanswer{
No.
}

\dsquestionex{Does the dataset contain data that might be considered sensitive in any way (e.g., data that reveals racial or ethnic origins, sexual orientations, religious beliefs, political opinions or union memberships, or locations; financial or health data; biometric or genetic data; forms of government identification, such as social security numbers; criminal history)?}{If so, please provide a description.}

\dsanswer{
No.
}

\dsquestion{Any other comments?}

\bigskip
\dssectionheader{Collection Process}

\dsquestionex{How was the data associated with each instance acquired?}{Was the data directly observable (e.g., raw text, movie ratings), reported by subjects (e.g., survey responses), or indirectly inferred/derived from other data (e.g., part-of-speech tags, model-based guesses for age or language)? If data was reported by subjects or indirectly inferred/derived from other data, was the data validated/verified? If so, please describe how.}

\dsanswer{
We performed in situ acquisitions at \censor{ESGT}. Annotation and 3D modeling were carried out by humans experts based on the acquired colorized point clouds, with some occasionnal in situ ground truth checks.

}

\dsquestionex{What mechanisms or procedures were used to collect the data (e.g., hardware apparatus or sensor, manual human curation, software program, software API)?}{How were these mechanisms or procedures validated?}

\dsanswer{
Data acquisition was carried out at \censor{ESGT} using two Terrestrial Laser Scanners (TLS): the RTC360 from Leica Geosystems (loaned by Leica Geosystems) and Trimble X7 from Trimble (loaned by \censor{ESGT}). High-resolution pictures are taken for each scan (almost 15MP for RTC360 and 10MP for Trimble X7). Scans are preregistered during the survey using respectively Cyclone Field on an iPad for RTC360 scans and with Realworks on a Trimble T10X for Trimble X7 scans.

We performed and bundled multiple scans inside every room to capture as many pieces of equipment as possible. Trimble X7 data is first registred on RealWorks and exported to e57 format. Subsequently, scans are imported into Register360 and merged with RTC360 data. During registration, any missing links are manually corrected. Then, the point cloud is georefenced using target coordinates obtained from Total Stations and GNSS.
}

\dsquestion{Who was involved in the data collection process (e.g., students, crowdworkers, contractors) and how were they compensated (e.g., how much were crowdworkers paid)?}

\dsanswer{The data collection process was carried out by the following persons:
\begin{itemize}
    \item \censor{Maxime Mérizette} (Ph.D. student): data collection, labeling quality checks, 3D CAD model quality check, point cloud processing (registration, exports),
    \item \censor{Lilian} (2nd year engineering student): data collection, point cloud processing (registration, exports), 3D CAD model creation,
    \item \censor{Léa Corduri, Judicaëlle Djeudji Tchaptchet, Damien Richard} (2nd year engineering students): point cloud labeling, class definition, comparison of annotation softwares. 
\end{itemize}
}

\dsquestionex{Over what timeframe was the data collected? Does this timeframe match the creation timeframe of the data associated with the instances (e.g., recent crawl of old news articles)?}{If not, please describe the timeframe in which the data associated with the instances was created.}

\dsanswer{RTC360 acquisition was carried out over three days (16 to 18 october 2023). Additional Trimble acquisitions were spread between late September and early November}

\dsquestionex{Were any ethical review processes conducted (e.g., by an institutional review board)?}{If so, please provide a description of these review processes, including the outcomes, as well as a link or other access point to any supporting documentation.}

\dsanswer{
No.
}

\dsquestionex{Does the dataset relate to people?}{If not, you may skip the remaining questions in this section.}

\dsanswer{
No, there are no identifiable persons in the dataset.
}

\bigskip
\dssectionheader{Preprocessing/cleaning/labeling}

\dsquestionex{Was any preprocessing/cleaning/labeling of the data done (e.g., discretization or bucketing, tokenization, part-of-speech tagging, SIFT feature extraction, removal of instances, processing of missing values)?}{If so, please provide a description. If not, you may skip the remainder of the questions in this section.}

\dsanswer{
The most obvious Lidar artefacts and outliers in the point clouds were cleaned and removed during the labeling process.
}

\dsquestionex{Was the “raw” data saved in addition to the preprocessed/cleaned/labeled data (e.g., to support unanticipated future uses)?}{If so, please provide a link or other access point to the “raw” data.}

\dsanswer{
Yes, raw point clouds are also available in the Bronze variant of the dataset.  
Manual semantic labeling was performed using the annotation tool available in 3DReshaper from Leica Geosytems.
}

\dsquestionex{Is the software used to preprocess/clean/label the instances available?}{If so, please provide a link or other access point.}

\dsanswer{
\begin{enumerate}
    \item \textbf{Preprocessing} \\
        Realworks \url{https://geospatial.trimble.com/fr/products/software/trimble-realworks} \\
        Register360 \url{https://leica-geosystems.com/fr-fr/products/laser-scanners/software/leica-cyclone/leica-cyclone-register-360}

    \item \textbf{Clean \& Label}: 3DReshaper \url{https://leica-geosystems.com/fr-fr/products/laser-scanners/software/leica-cyclone/leica-cyclone-3dr}
\end{enumerate}

}

\dsquestion{Any other comments?}

\dsanswer{
}

\bigskip
\dssectionheader{Uses}

\dsquestionex{Has the dataset been used for any tasks already?}{If so, please provide a description.}

\dsanswer{
Yes, initial baselines for semantic segmentation of point clouds have been tried on the 3DSES dataset using deep models for point cloud segmentation (\ie Swin3D \cite{yang_swin3d_2023}).
}

\dsquestionex{Is there a repository that links to any or all papers or systems that use the dataset?}{If so, please provide a link or other access point.}

\dsanswer{
Not currently.
}

\dsquestion{What (other) tasks could the dataset be used for?}

\dsanswer{In addition to the task of semantic segmentation, the dataset could also be used for:
\begin{itemize}
    \item unsupervised pretraining of deep models on point clouds,
    \item point cloud colorization,
    \item scan-to-BIM, \ie extracting a (semantic) 3D CAD model from a point cloud,
    \item novel view generation,
    \item automated labeling of point clouds based on 3D CAD models.
\end{itemize}
}

\dsquestionex{Is there anything about the composition of the dataset or the way it was collected and preprocessed/cleaned/labeled that might impact future uses?}{For example, is there anything that a future user might need to know to avoid uses that could result in unfair treatment of individuals or groups (e.g., stereotyping, quality of service issues) or other undesirable harms (e.g., financial harms, legal risks) If so, please provide a description. Is there anything a future user could do to mitigate these undesirable harms?}

\dsanswer{
The 3DSES dataset uses acquisitions from specific sensors: Leica RTC360 and Trimble X7. These sensors have particuliar characteristics that might not be representative of future sensors, especially regarding the calibration of radiometric features (\ie intensity). Findings on this dataset might not necessarily transfer exactly on other point cloud datasets acquired by different sensors, especially datasets not using TLS sensors.
}

\dsquestionex{Are there tasks for which the dataset should not be used?}{If so, please provide a description.}

\dsanswer{
To the best of our knowledge, no.
}

\dsquestion{Any other comments?}

\bigskip
\dssectionheader{Distribution}

\dsquestionex{Will the dataset be distributed to third parties outside of the entity (e.g., company, institution, organization) on behalf of which the dataset was created?}{If so, please provide a description.}

\dsanswer{
No.
}

\dsquestionex{How will the dataset will be distributed (e.g., tarball on website, API, GitHub)}{Does the dataset have a digital object identifier (DOI)?}

\dsanswer{
The dataset will be made available on a public archive (\eg Zenodo). A competition will also be hosted on Codabench.
}

\dsquestion{When will the dataset be distributed?}

\dsanswer{Circa November 2024.
}

\dsquestionex{Will the dataset be distributed under a copyright or other intellectual property (IP) license, and/or under applicable terms of use (ToU)?}{If so, please describe this license and/or ToU, and provide a link or other access point to, or otherwise reproduce, any relevant licensing terms or ToU, as well as any fees associated with these restrictions.}

\dsanswer{The dataset will be distributed under the \href{https://creativecommons.org/licenses/by-sa/4.0/}{Creative Commons CC BY-SA 4.0} license. It permits free access and use of the dataset, alongside redistribution and adaptation under the same terms, provided attribution is given.
}

\dsquestionex{Have any third parties imposed IP-based or other restrictions on the data associated with the instances?}{If so, please describe these restrictions, and provide a link or other access point to, or otherwise reproduce, any relevant licensing terms, as well as any fees associated with these restrictions.}

\dsanswer{No.
}

\dsquestionex{Do any export controls or other regulatory restrictions apply to the dataset or to individual instances?}{If so, please describe these restrictions, and provide a link or other access point to, or otherwise reproduce, any supporting documentation.}

\dsanswer{No.
}

\dsquestion{Any other comments?}

\dsanswer{
}

\bigskip
\dssectionheader{Maintenance}

\dsquestion{Who will be supporting/hosting/maintaining the dataset?}

\dsanswer{The dataset will be supported and maintained by \censor{GeF laboratory from ESGT}. Hosting will be provided graciously by Zenodo. Reference code and information will be hosted on a GitHub repository.
}

\dsquestion{How can the owner/curator/manager of the dataset be contacted (e.g., email address)?}

\dsanswer{
The main author can be contacted by email: \censor{\texttt{maxime.merizette@lecnam.net}}.
Alternatively, the \censor{lab director} can be contacted at \censor{\texttt{jerome.verdun@cnam.fr}}.
}

\dsquestionex{Is there an erratum?}{If so, please provide a link or other access point.}

\dsanswer{Not currently.
}

\dsquestionex{Will the dataset be updated (e.g., to correct labeling errors, add new instances, delete instances)?}{If so, please describe how often, by whom, and how updates will be communicated to users (e.g., mailing list, GitHub)?}

\dsanswer{
Yes, the dataset might eventually be updated to cover the entirety of \censor{ESGT} building and/or to include other modalities, such as panoramic pictures.

Errors due to labeling will be able to be raised on GitHub and might be corrected depending on their gravity.

All updates will be published on GitHub and Zenodo.
}

\dsquestionex{If the dataset relates to people, are there applicable limits on the retention of the data associated with the instances (e.g., were individuals in question told that their data would be retained for a fixed period of time and then deleted)?}{If so, please describe these limits and explain how they will be enforced.}

\dsanswer{
Not applicable.
}

\dsquestionex{Will older versions of the dataset continue to be supported/hosted/maintained?}{If so, please describe how. If not, please describe how its obsolescence will be communicated to users.}

\dsanswer{
Obsolescence of older versions will be communicated to users using the same channels as update announcements.
}

\dsquestionex{If others want to extend/augment/build on/contribute to the dataset, is there a mechanism for them to do so?}{If so, please provide a description. Will these contributions be validated/verified? If so, please describe how. If not, why not? Is there a process for communicating/distributing these contributions to other users? If so, please provide a description.}

\dsanswer{
Willing contributors will be able to manifest their interest on GitHub.
}

\dsquestion{Any other comments?}

\dsanswer{
}